\documentclass[runningheads]{llncs}
\usepackage{graphicx}
\usepackage{amsmath,amssymb} 
\usepackage{color}
\usepackage{subfigure}
\usepackage{caption}
\usepackage{multirow}
\usepackage{wrapfig}
\usepackage{marvosym}
\usepackage{ifsym}

\begin{document}
\pagestyle{headings}
\mainmatter


\title{Network Pruning via Feature Shift Minimization} 
\titlerunning{}
\authorrunning{ }

\author{Yuanzhi Duan\inst{1} \and
Yue Zhou\inst{1} \and 
Peng He\inst{1} \and 
Qiang Liu\inst{2} \and 
Shukai Duan\inst{1}  \and \\
Xiaofang Hu\inst{1}\thanks{Corresponding author}\orcidID{0000-0003-3764-2640}
}
\institute{College of Artificial Intelligence, Southwest University, Chongqing, China \and
Harbin Institute of Technology, Harbin, China \\
\email{\{huxf,duansk\}@swu.edu.cn, \{swuyzhid,hepeng5\}@email.swu.edu.cn, zhouyuenju@163.com, 18b933041@stu.hit.edu.cn}}
\maketitle

\begin{abstract}
Channel pruning is widely used to reduce the complexity of deep network models.
Recent pruning methods usually identify which parts of the network to discard by proposing a channel importance criterion.
However, recent studies have shown that these criteria do not work well in all conditions.
In this paper, we propose a novel Feature Shift Minimization (FSM) method to compress CNN models, which evaluates the feature shift by converging the information of both features and filters.
Specifically, we first investigate the compression efficiency with some prevalent methods in different layer-depths and then propose the feature shift concept.
Then, we introduce an approximation method to estimate the magnitude of the feature shift, since it is difficult to compute it directly.
Besides, we present a distribution-optimization algorithm to compensate for the accuracy loss and improve the network compression efficiency.
The proposed method yields state-of-the-art performance on various benchmark networks and datasets, verified by extensive experiments.
Our codes are available at: \url{https://github.com/lscgx/FSM}.
\end{abstract}

\section{Introduction}

The rapid development of convolutional neural networks (CNN) has obtained remarkable success in a wide range of computer vision applications, such as image classification \cite{He2016DeepRL,Huang2017DenselyCC,Krizhevsky2012ImageNetCW}, video analysis \cite{Girdhar2019DistInitLV,Jiang2018DeepVSAD,Lin2019BMNBN}, object detection \cite{Girshick2014RichFH,Guo2020HitDetectorHT,Ren2015FasterRT}, semantic segmentation \cite{Chen2018DeepLabSI,Chen2018EncoderDecoderWA,Shelhamer2017FullyCN}, etc.
The combination of CNN models and IoT devices yields significant economic and social benefits in the real world.
However, better performance for a CNN model usually means higher computational complexity and a greater number of parameters, limiting its application in resource-constrained devices.
Therefore, model compression techniques are required.

To this end, compressing the existing network is a popular strategy, including tensor decomposition \cite{Raja2019ObtainingSI}, parameter quantification \cite{Liu2020ReActNetTP}, weight pruning \cite{Han2015LearningBW,Li2017PruningFF}, structural pruning \cite{Lin2020HRankFP}, etc.
Moreover, another strategy is to create a new small network directly, including knowledge distillation \cite{Hinton2015DistillingTK} and compact network design \cite{Howard2017MobileNetsEC}.
Among these compression techniques, structural pruning has significant performance and is usable for various network architectures.
In this paper, we present a channel pruning method, belonging to structural pruning, for model compression.

\begin{figure}[t]
\centering
\begin{minipage}[!t]{1.\linewidth}
\centering
\subfigure[VGGNet-1]{
	\centering
	\includegraphics[scale=0.13]{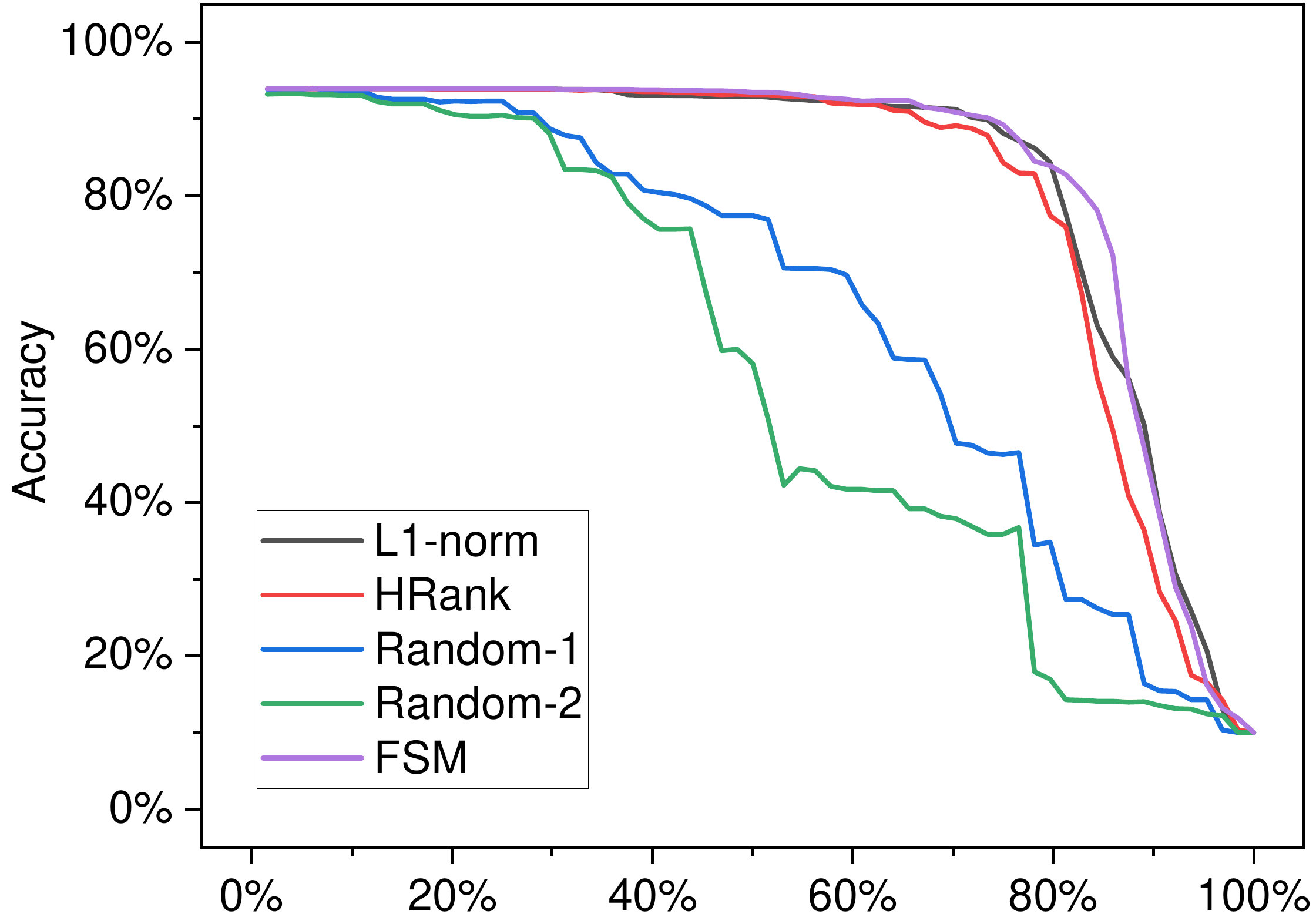}}
\subfigure[VGGNet-6]{
	\centering
	\includegraphics[scale=0.13]{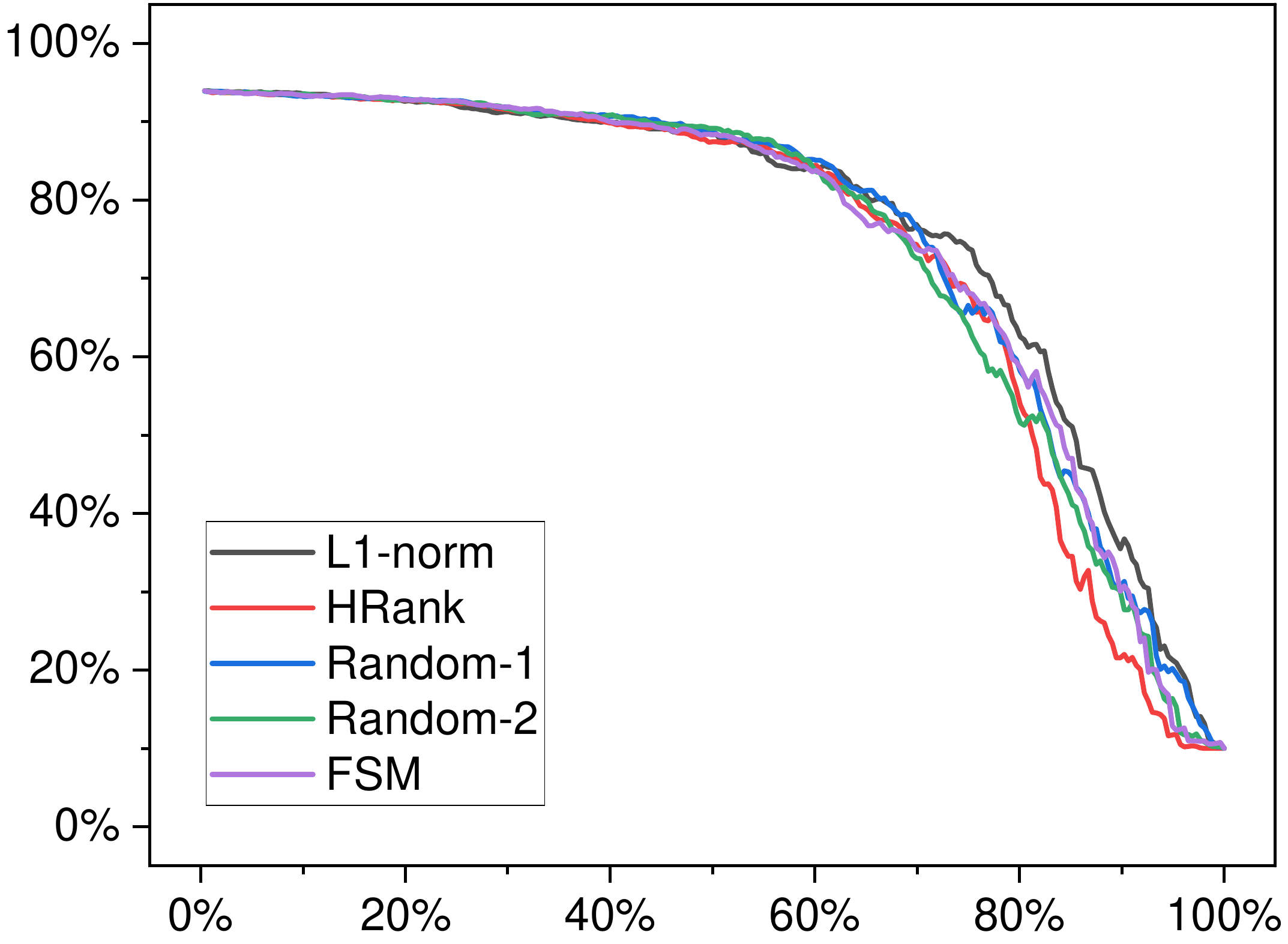}}
\subfigure[VGGNet-11]{
	\centering
	\includegraphics[scale=0.13]{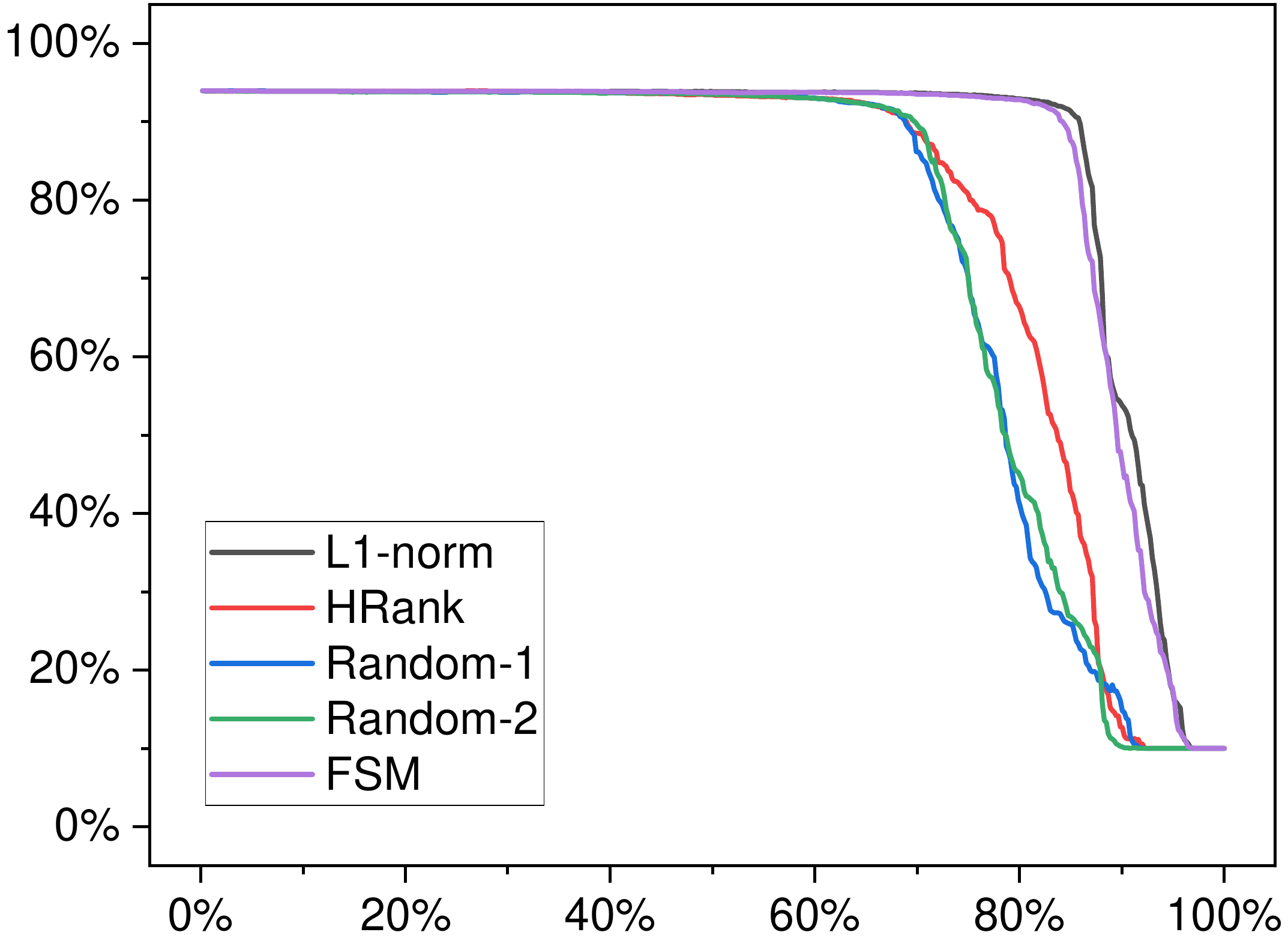}}
\subfigure[ResNet-56-1]{
	\centering
	\includegraphics[scale=0.13]{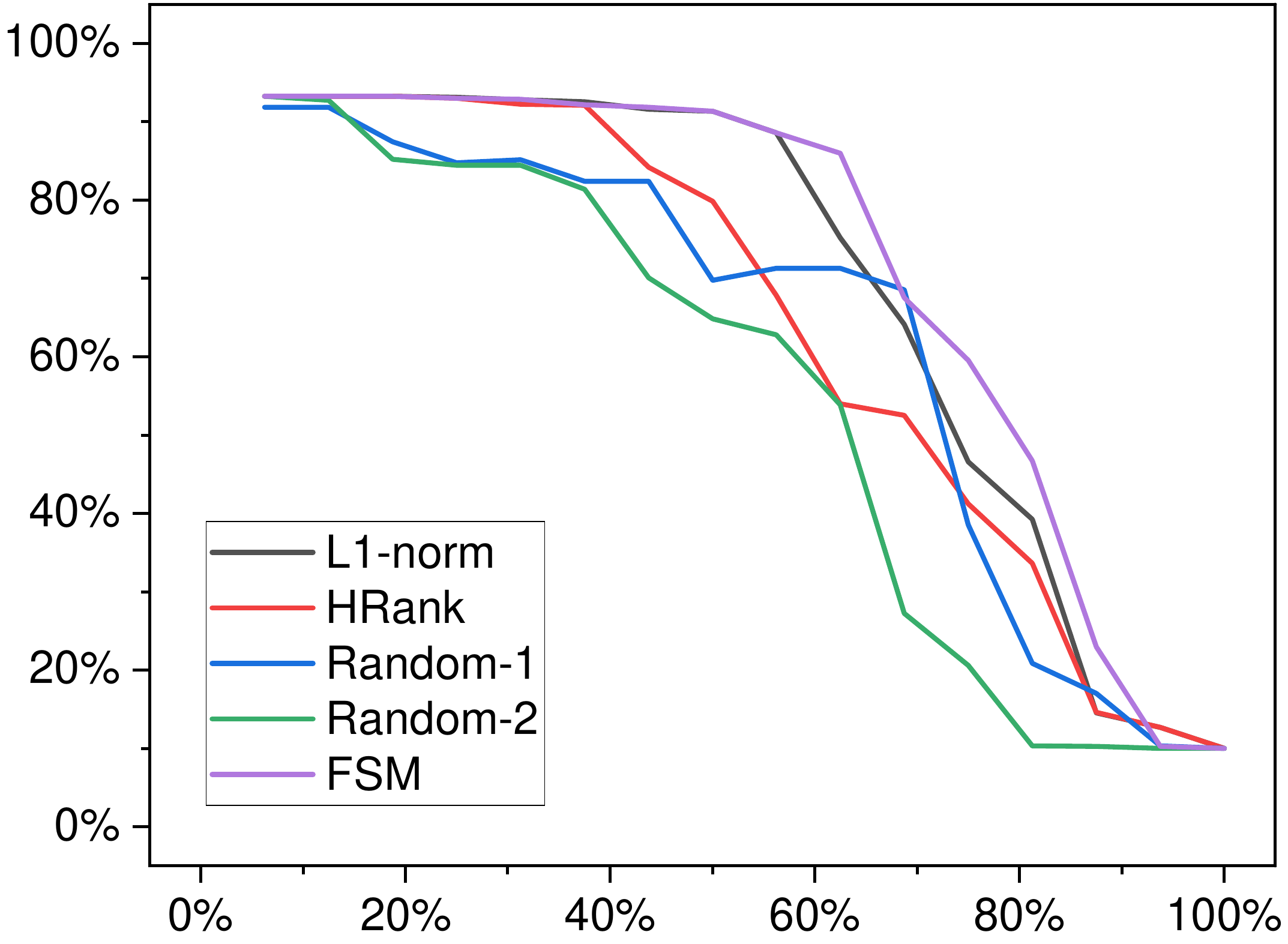}}
\end{minipage}
\begin{minipage}[!t]{1.0\linewidth}
\centering
\subfigure[ResNet-50-1]{
	\centering
	\includegraphics[scale=0.13]{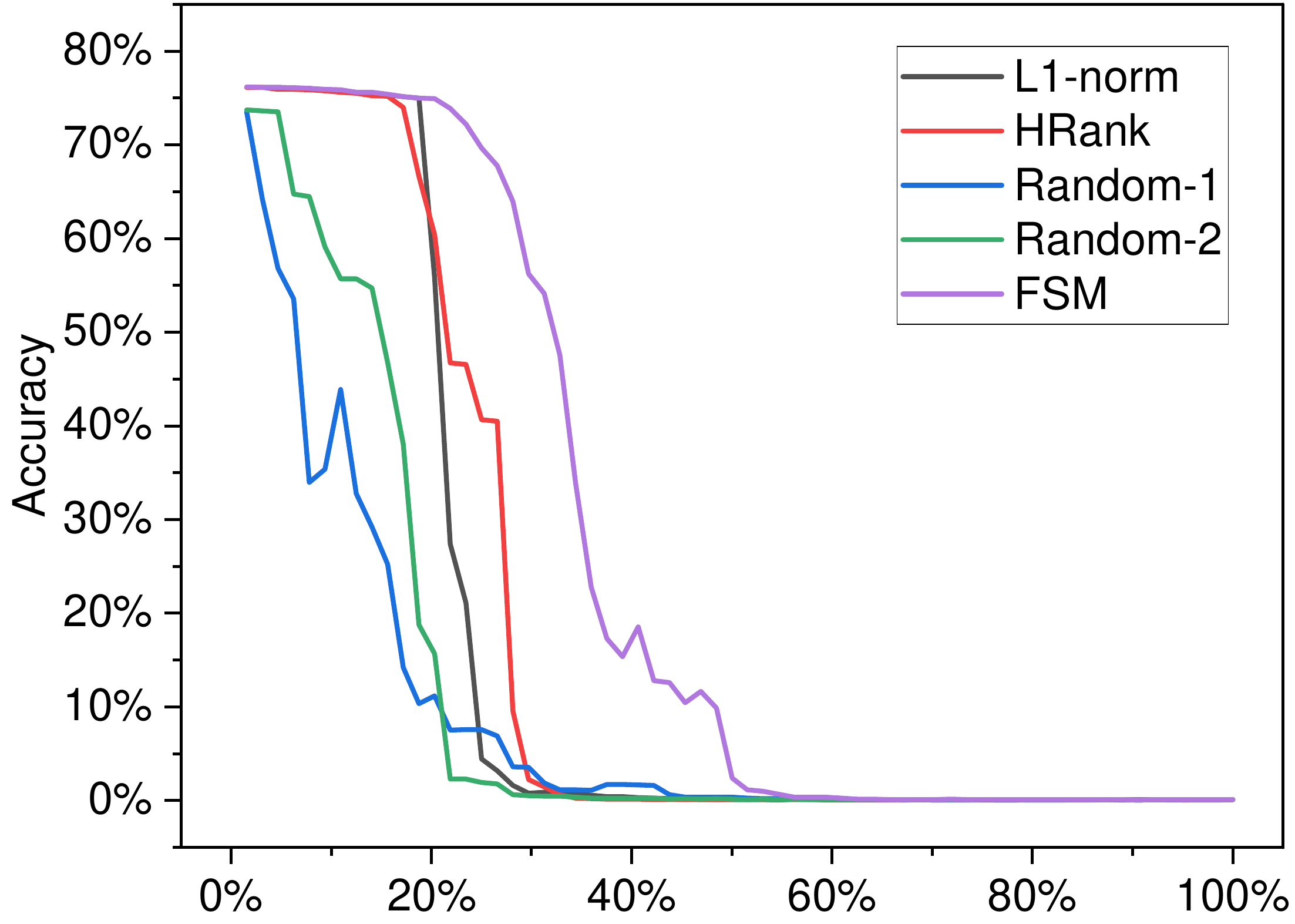}}
\subfigure[ResNet-50-11]{
	\centering
	\includegraphics[scale=0.13]{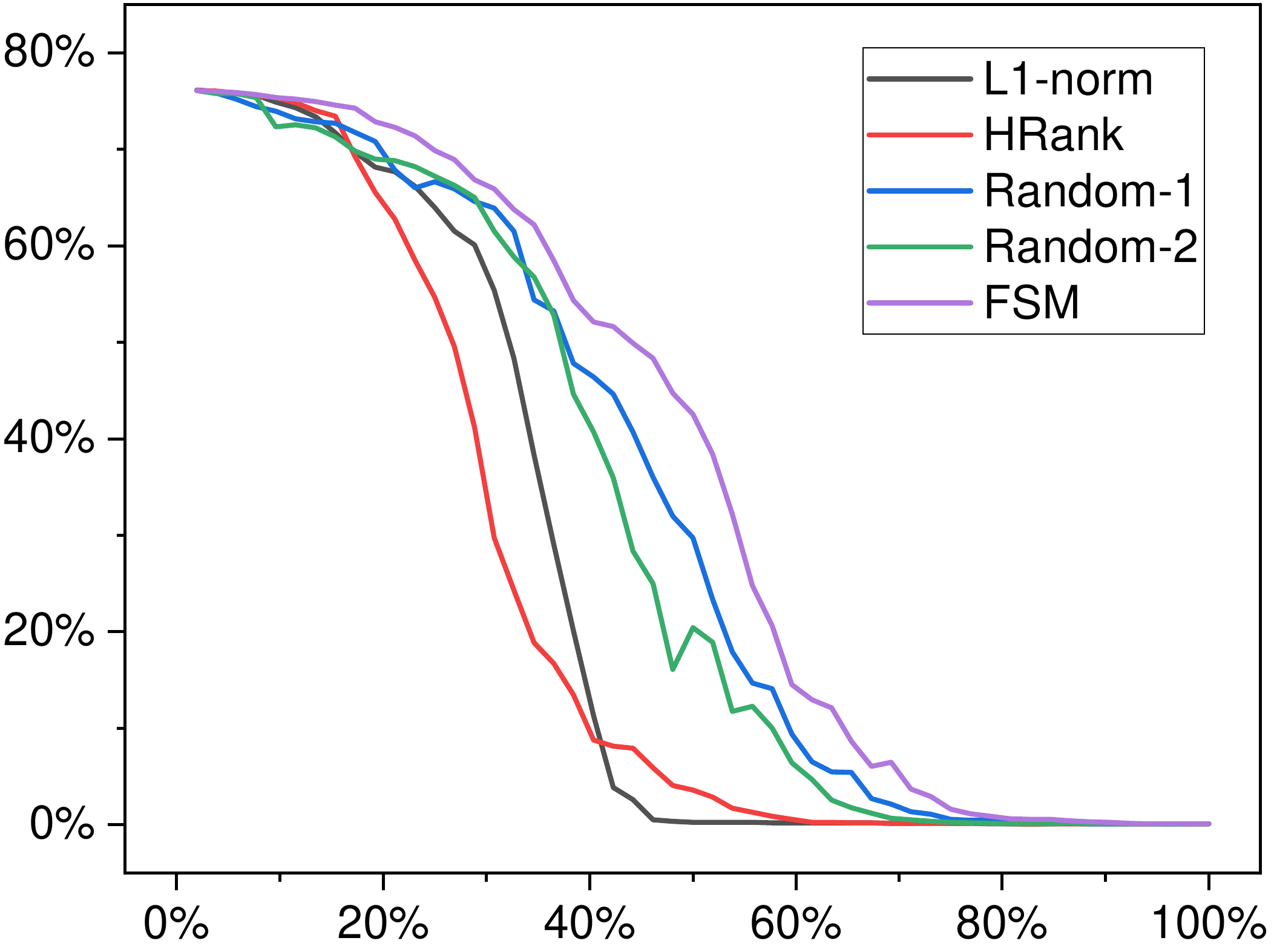}}
\subfigure[ResNet-50-23]{
	\centering
	\includegraphics[scale=0.13]{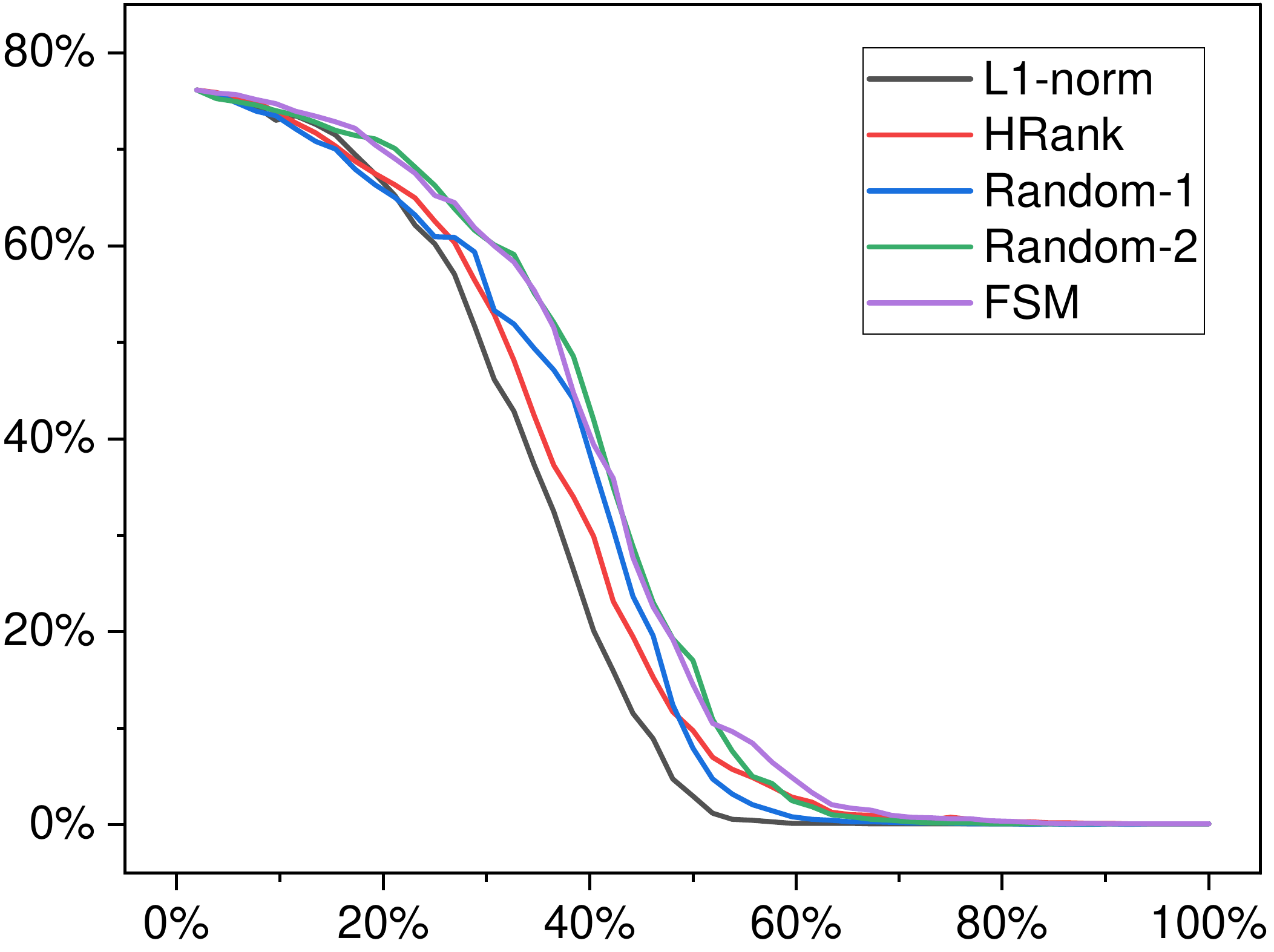}}
\subfigure[GoogLeNet-1]{
	\centering
	\includegraphics[scale=0.13]{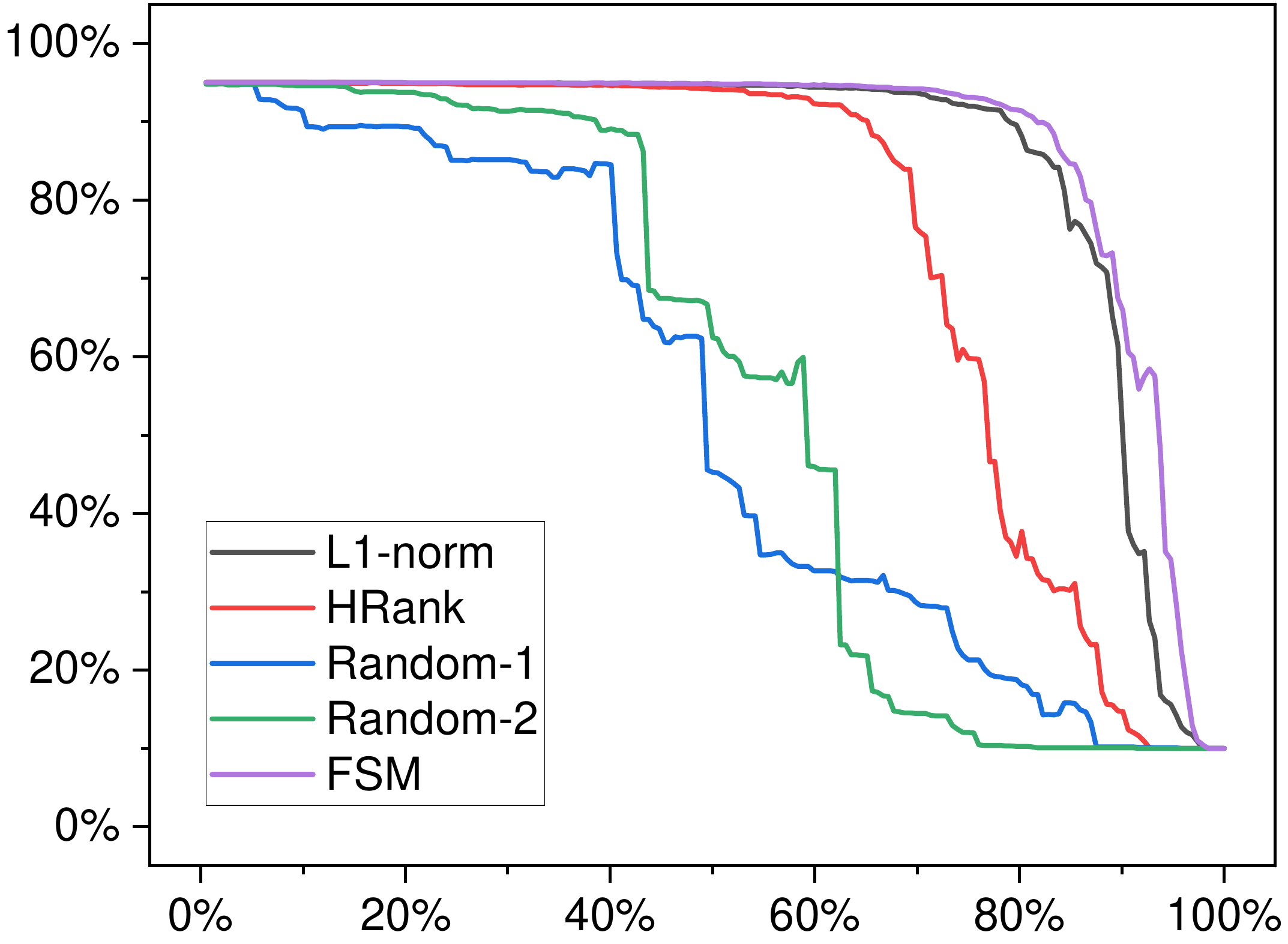}}
\end{minipage}
\caption{Analysis of the relationship between compression rate, layer depth, accuracy, and methods. Names such as "ResNet-50-1" mean experiment on the 1st layer of ResNet-50. For each subﬁgure, the horizontal axis is the compression rate, and the vertical axis is the accuracy.}
\label{fig:analysis}
\end{figure}

In the process of channel pruning, some channels that are considered unimportant or redundant are discarded and get a sub-network of the original network. 
The core of channel pruning lies in the design of filter or feature selection criteria.
One idea is to directly use the inherent properties of filters or features, such as statistical properties, matrix properties, etc.
For example, Li \emph{et al.} \cite{He2018PruningFV,Li2017PruningFF} sorting the filters by $l_1$ norm of filters, and Lin \emph{et al.} \cite{Lin2020HRankFP} prune channels by calculating the rank of features.
Another idea is to evaluate the impact of removing a channel on the next layer or the accuracy loss \cite{Luo2017ThiNetAF}.
These criteria, which are based on different properties, succeeded in model compression, but the performance faded in some conditions.
Because it is difficult for a single criterion to take into account all factors such as feature size or feature dimension at the same time.
So, many pruning criteria are suboptimal because they only work well locally, not globally. 
As shown in Fig.\;\ref{fig:analysis}, in some cases, random pruning can even outperform well-designed methods.

To compensate for the limitations of a single criterion, some multi-criteria approaches have been proposed recently.
For example, in \cite{Li2021TowardsCC}, a collaborative compression method was proposed for channel pruning, which uses the information of compression sensitivity for each layer. 
Similarly, in \cite{Wang2021ConvolutionalNN}, the structural redundancy information of each layer is used to guide pruning.
Although these methods take into account the influence of other factors on pruning efficiency, they make no improvements to the criteria and require additional computational consumption.

\begin{figure}[t]
\centering
\includegraphics[scale=0.35]{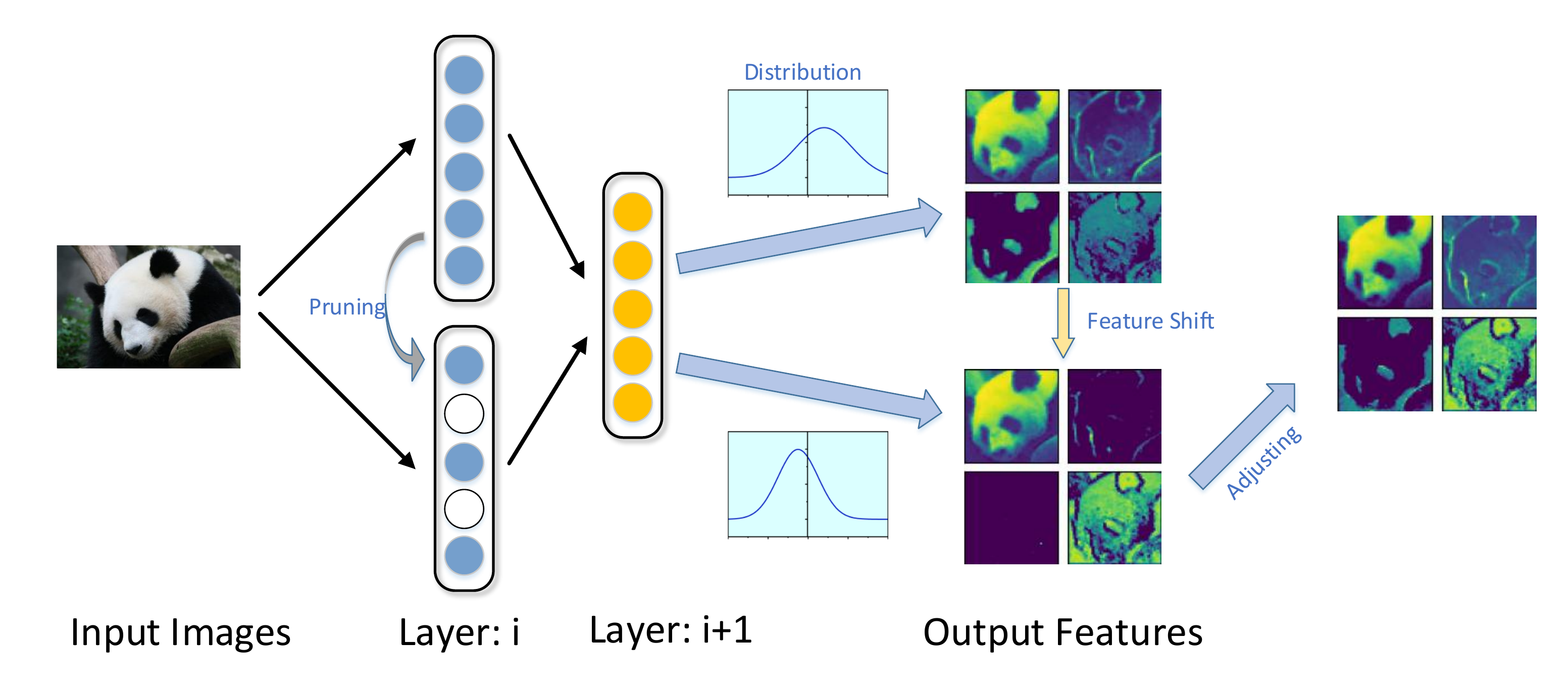}
\caption{Diagram of the proposed Feature Shift Minimization method (FSM).}
\label{fig:process}
\end{figure}

In this paper, we propose a new channel pruning method by minimizing feature shift.
We define feature shift as the changes in the distribution of features in a layer due to pruning.
As shown in Fig.\;\ref{fig:process}, the feature details changed or disappeared due to pruning, but when the distribution of the features is properly adjusted, the disappeared feature details emerge again.
This experiment was done on ImageNet, using the ResNet-50 model.
This suggests that the loss of feature detail, caused by pruning some channels, may not have really disappeared, which is related to the feature activation state.
Even those channels that are considered unimportant may cause changes in the distribution of features, which in turn lead to under-activation or over-activation of features. 
Therefore, it is reasonable to minimize the feature shift for model compression.
Moreover, we mathematically demonstrate that feature shift is an important factor for channel pruning and analyze the effect of the activation functions, in Section\;\ref{Filter Pruning Analysis}.

Computing feature shift directly is not an easy task, as it requires traversing the entire training dataset. 
To address this issue, we propose an evaluation method to measure the feature shift, which combines information from both filters and features.
Based on this, we compress models by evaluating the feature shift when each channel is removed, which performs well in different feature dimensions.
In particular, the proposed pruning method does not require sampling the features and uses only the parameters of the pre-trained models.
In addition, to prove that the feature shift occurs in the pruning process, we design a distribution-optimization algorithm to adjust the pruned feature distribution, which significantly recovers the accuracy loss, as will be discussed in Section\;\ref{Distribution Optimization}.

To summarize, our main contributions are as follows:
\begin{itemize}
  \item[1)]We investigate the relationship between compression rate, layer depth, and model accuracy with different pruning methods. It reveals that the feature shift is an important factor affecting pruning efficiency. 
  \item[2)]We propose a novel channel pruning method, Feature Shift Minimization (FSM), which combines information from both features and filters. Moreover, a distribution-optimization algorithm is designed to accelerate network compression.
  \item[3)]Extensive experiments on CIFAR-10 and ImageNet, using VGGNet, MobileNet, GoogLeNet, and ResNet, demonstrate the effectiveness of the proposed FSM.
\end{itemize}

\section{Related Work}\label{Related Work} 
\textbf{Channel pruning.}
Channel pruning has been successfully applied to the compression of various network architectures, which lies in the selection of filters or features.
Some previous methods use the intrinsic properties of filters to compress models.
For example, \cite{He2018PruningFV,Li2017PruningFF} prune filters according to their $l_i$ norms. 
In \cite{Lin2020HRankFP}, the rank of the features is used to measure the information richness of the features and retain the features with high information content.
Some other approaches go beyond the limits of a single layer.
For example, Chin \emph{et al.} \cite{Chin2020TowardsEM} removes filters by calculating the global ranking.
Liu \emph{et al.}\cite{Liu2017LearningEC} analyzes the relationship between the filters and BN layers and designs a scale factor to represent the importance.
In addition, computing the statistical information of the next layer and minimizing the feature reconstruction error is also a popular idea \cite{He2017ChannelPF,Luo2017ThiNetAF}.
Different from methods that are based on only a single filter or feature, \cite{Singh2020LeveragingFC,Wang2021ConvolutionalNN,Wang2021ModelPB} remove redundant parts of a model by measuring the difference between the filters or features. 

\noindent \textbf{Effect of Layer-Depth.}
For different layers, the feature size, dimensions, and redundancy are different.
Recent works have found that different layers of a model have different sensitivities to the compression rate.
Wang \emph{et al.} \cite{Wang2021ConvolutionalNN} proposes that the structural redundancy of each layer are different, which plays a key role in the pruning process. 
In layers with high structural redundancy, more filters can be safely discarded with little loss of accuracy.
Li \emph{et al.} \cite{Li2021TowardsCC} represents that the compression sensitivity of each layer is distinct, and it is used to guide pruning.
In addition, He \emph{et al.} \cite{He2020LearningFP} propose that a pruning criterion does not always work well in all layers, and they achieve better performance by using different pruning criteria on different layer-depths.
Earlier methods \cite{He2018SoftFP,Huang2018DataDrivenSS,Zhuang2018DiscriminationawareCP} required iterative tuning of the compression ratio to achieve better model accuracy. It is tedious and time-consuming.
Considering the difference between different layers can greatly improve the pruning efficiency.

We investigate the trend of accuracy when adopting different compression rates in different layer-depths.
Two methods (L1 and HRank) are investigated, which sort filters and features, respectively. 
Based on this, we propose a new perspective to explore the redundancy of neural networks.

\section{Methodology}\label{Methodology}
In this section, we analyze the process of filter pruning and point out that the distribution of features is changed after pruning.
Then we propose a novel pruning criterion that minimizes the feature shift.
Unlike some heuristic pruning criteria, there is much theoretical analysis to support the proposed pruning algorithm.
Moreover, a distribution optimization algorithm is presented to restore the model accuracy.

\subsection{Filter Pruning Analysis}\label{Filter Pruning Analysis}
Filter pruning aims to remove those relatively unimportant filters, which evaluate the importance of the filter or feature by a specific importance criterion. However, we found that these criteria do not work well in all layers. We conduct experiments on CIFAR-10 with $l1$-norm and HRank, which sort filters and features, respectively. In addition, we also test the performance of random pruning in different layers. As shown in Fig.\;\ref{fig:analysis}, two phenomena are obvious: a)As the compression rate increases, the drop in accuracy becomes more obvious. When the compression rate exceeds a critical value, the accuracy declines dramatically. b) For high-dimensional features, sorting filters or feature maps by these criteria isn't substantially better than random pruning.

The phenomena reveal that: a) Exist other factors damage the accuracy, and their impact grows with the compression rate. 
b) For low-dimensional features, pruning criteria are effective. However, for high-dimensional features, random pruning also performs well. This indicates that the gap between the importance scores of filters/features becomes smaller as the dimensionality increases.

To clarify why these phenomena occur, we mathematically analyze the changes that occur in one feature map before and after pruning.
A typical inference process in one layer is Input-Conv-BN\cite{Ioffe2015BatchNA}-ReLU\cite{Glorot2011DeepSR}-Output. 

Many benchmark network architectures, such as ResNet, DenseNet, and GoogLeNet, are based on it.
In particular, we let $X_{i}=(x^{(1)}_{i}\cdots x^{(d_{i})}_{i})$ represent the input for $i$-th layer of one CNN model, where $d_{i}$ stand for the number of the dimensions. 
Firstly, we normalize each dimension on BN layer  
\begin{align}\label{eq1}
  \widehat{x}^{(k)}_{i} =\frac{x^{(k)}_{i}-E[x^{(k)}_{i}]}{\sqrt{Var[x^{(k)}_{i}]}},\, y^{(k)}_{i}=\gamma^{(k)}_{i}\widehat{x}^{(k)}_{i} +\beta^{(k)}_{i}, 
\end{align}
\noindent where $E$ is the expectation, $Var$ is the variance, and $k$ stand for the $k$-th dimension.
We let $Y_{i}=(y^{(1)}_{i}\cdots y^{(d_{i})}_{i})$ denotes the output of the BN layer.
The output of the $k$-th dimension has a mean of $\beta^{(k)}_{i}$ and a standard deviation of $\gamma^{(k)}_{i}$, which can be directly obtained by the pre-trained models.
Then, $Y_{i}$ is passed on to the ReLU layer.
For each of the values in $Y_{i}$, the ReLU activation function is applied to it to get thresholded values. The output can be formulated as:
\begin{align}
\widehat{y}^{(k)}_{i}=ReLU(y^{(k)}_{i})=max(0,y^{(k)}_{i}).
\end{align}
Obviously, the distribution of $\widehat{y}^{(k)}_{i}$ changes after passing through the ReLU layer.
The output expectation
\begin{align}
E[\widehat{y}^{(k)}_{i}]=E[max(0,y^{(k)}_{i})],
\end{align}
and the input of the $(i+1)$-th layer
\begin{align}
x^{(k)}_{i+1} =(\widehat{y}^{(1)}_{i}\cdots \widehat{y}^{(d_{i})}_{i})\cdot w_{i+1} \ne (\widehat{y}^{(1)}_{i}\cdots \widehat{y}^{(\widehat{d}_{i})}_{i})\cdot w_{i+1},\, s.t.\,\, \widehat{d}_{i}\leq {d}_{i},
\end{align}
where $w$ represents the weights and $\widehat{d}_{i}$ stands for the $i$-th dimension after pruning. 
As can be seen, the distribution of input $x$ of one channel will shift when pruning some channels in the previous layer.
We assume that the shifted expectation of $x^{(k)}_{i}$ as $\widehat{E}[x^{(k)}_{i}]$, and the shifted variance as $\widehat{Var}[x^{(k)}_{i}]$. 
More, we let $\Delta_{E}=E[x^{(k)}_{i}]-\widehat{E}[x^{(k)}_{i}]$.
So we can reformulate the Eq.\,(\ref{eq1}) as:
\begin{align}
y^{(k)}_{i}&=\gamma^{(k)}_{i}\widehat{x}^{(k)}_{i} +\beta^{(k)}_{i} 
=\gamma^{(k)}_{i}\frac{x^{(k)}_{i}-E[x^{(k)}_{i}]}{\sqrt{Var[x^{(k)}_{i}]}} +\beta^{(k)}_{i} \\
&=\gamma^{(k)}_{i}\frac{x^{(k)}_{i}-(\widehat{E}[x^{(k)}_{i}]+\Delta_{E})}{\sqrt{Var[x^{(k)}_{i}]}} +\beta^{(k)}_{i} \\
&=\frac{\sqrt{\widehat{Var}[x^{(k)}_{i}]}}{\sqrt{Var[x^{(k)}_{i}]}}\gamma^{(k)}_{i}\frac{x^{(k)}_{i}-\widehat{E}[x^{(k)}_{i}]}{\sqrt{\widehat{Var}[x^{(k)}_{i}]}} +\beta^{(k)}_{i} - \gamma^{(k)}_{i}\frac{\Delta_{E}}{\sqrt{Var[x^{(k)}_{i}]}} \label{eq5}
\end{align}
The mean of $y^{(k)}_{i}$ is $\beta^{(k)}_{i} - \gamma^{(k)}_{i}\frac{\Delta_{E}}{\sqrt{Var[x^{(k)}_{i}]}}$, and the standard deviation shifted to $\frac{\sqrt{\widehat{Var}[x^{(k)}_{i}]}}{\sqrt{Var[x^{(k)}_{i}]}}\gamma^{(k)}_{i}$.

As a result, the expectation $E[\widehat{y}^{(k)}_{i}]$ is shifted after pruning, which directly affects the input to the next layer. At the same time, due to the ReLU activation function, some values that were not activated before being activated, or the activated values are deactivated.
We refer to the change in the distributions after the ReLU layer in the process of pruning as {\it feature shift}.

\setlength\tabcolsep{4pt}
\setlength{\intextsep}{0pt}
\begin{wraptable}{r}{0.5\linewidth}
  \centering
  \caption{Experiments on ResNet-50-23. FSM-R means pruning in reverse order.}
  \label{table:feature shift}
  \begin{tabular}{c|c|c|c}
    \hline\ 
    Rate        & 20\%   & 30\%   & 40\%      \\
    \hline\ 
    FSM(\%)     & 70.42  & 56.50 & 32.20  \\
    \hline\ 
    FSM-R(\%)   & 32.89  & 8.128  & 1.62   \\
    \hline
  \end{tabular}
\end{wraptable}

Predictably, at high compression rates, the magnitude of the feature shift may be greater compared to low rates, and the drop in accuracy may also be greater.
To prove it, we consider the feature shift of each channel as the selection criterion in pruning. We prune channels by their expectation of the feature shift. The channels with the least shift are discarded first.
As shown in Fig.\;\ref{fig:analysis}, many experiments show that the decreasing trend of the accuracy of the proposed method is less pronounced than other prevalent methods.
Our method can maintain higher accuracy compared to others at the same compression rate (e.g., 11.3\%/L1 vs. 8.79\%/HRank vs. 66.52\%/FSM with a ratio 40\% on ResNet-50-11).
It reveals that the feature shift is one of the critical factors for performance degradation.
In particular, when we prune the model in reverse order, the accuracy catastrophically decreases, as shown in Table~\ref{table:feature shift}.
It demonstrates that the greater the magnitude of the feature shift, the greater the loss of accuracy. The decrease in accuracy has a positive correlation with the feature shift.
So, it is reasonable to use the feature shift to guide pruning.
Features that exhibit less feature shift are relatively unimportant.

\subsection{Evaluating the Feature Shift}\label{Evaluating the feature shift}
As demonstrated in Eq.\,(\ref{eq5}), we need to calculate $\widehat{E}[x^{(k)}_{i}]$, which stands for the expectation of the $k$-th dimension in the $i$-th layer.
It is difficult and time-consuming to calculate it directly over the entire training dataset.
We present an approximation method to estimate the expectation of features and prove the feasibility in Section\;\ref{Error of the Feature Shift Evaluation}.
Specifically, we first expand the $\widehat{E}[x^{(k)}_{i}]$ as:
\begin{align}\label{eq9}
\widehat{E}[x^{(k)}_{i}]&=\widehat{E}[(\widehat{y}^{(1)}_{i-1}\cdots\widehat{y}^{(\widehat{d}_{i-1})}_{i-1})\cdot w_{i}^{(k)}] \\
&=\widehat{E}[\widehat{y}^{(1)}_{i-1}]*w_{i}^{(k,1)}+\cdots+\widehat{E}[\widehat{y}^{(\widehat{d}_{i-1})}_{i-1}]*w_{i}^{(k,\widehat{d}_{i-1})},
\end{align}
where $\widehat{d}_{i-1}$ stands for the $(i-1)$-th dimension after pruning, and $\widehat{d}_{i-1}\leq{d}_{i-1}$.
Moreover, according to Eq.\,(\ref{eq1}), we get
\begin{align}\label{eq10}
\widehat{E}[\widehat{y}^{(k)}_{i-1}]&=E[max(0,y^{(k)}_{i-1})] 
\approx \int_{0}^{\infty} x \frac{1}{\gamma^{(k)}_{i-1}\sqrt{(2\pi)}}e^{-\frac{(x-\beta^{(k)}_{i-1})^2}{2(\gamma^{(k)}_{i-1})^{2}}}dx.
\end{align}
Without computing the expectation over the entire train dataset, the approximation of $\widehat{E}[x^{(k)}_{i}]$ can be obtained by combining Eq.\,(\ref{eq9}) and Eq.\,(\ref{eq10}).
Then, according to Eq.\,(\ref{eq5}), we get the shifted expectation and the standard deviation of the features.
\subsection{Filter Selection Strategy}\label{Filter Selection Strategy}
The pruning strategy of a layer can be regarded as minimizing the sum of the feature shift of all channels of the pruned model.
For a CNN model with $N$ layers, the optimal pruning strategy can be expressed as an optimization problem:
\begin{align}
arg \, min\sum\limits_{i=1}^{N}\sum\limits_{k=1}^{\widehat{d_i}}\left| \Delta_{E^{(k)}_{i}}\right|=arg \, min\sum\limits_{i=1}^{N}\sum\limits_{k=1}^{\widehat{d_i}}\left| E[x^{(k)}_{i}]-\widehat{E}[x^{(k)}_{i}]\right|,
\end{align}
where $\widehat{d_i}$ is the dimension of each layer after pruning.
During the pruning of a layer, we evaluate the effect of each feature for the feature shift of the next layer, as follows:
\begin{align}\label{eq15}
\delta(x^{(k)}_{i})=\sum\limits_{j=1}^{d_{i+1}}\left| w_{i+1}^{(j,k)}*E[y^{(k)}_{i}]\right|,
\end{align}
where $w_{i+1}^{(j,k)}$ denotes the $k$-th vector of the $(i+1)$-th layer's $j$-th dimension.
$\delta(x^{(k)}_{i})$ represents the sum of feature shift of one output feature to all channels in the $(i+1)$-th layer. 
Then, we sort all features according to the values $\delta(\cdot)$. 
Features with low values are considered unimportant and will be discarded in preference.

\subsection{Distribution Optimization}\label{Distribution Optimization}
For a pruned model, we present a simple distribution optimization method to recover accuracy. 
As presented in Eq.\,(\ref{eq5}), the distribution of $y^{(k)}_{i}$ was changed due to the shift of   $E[x^{(k)}_{i}]$ and $Var[x^{(k)}_{i}]$ after pruning. 
Hence, we can adjust them to $\widehat{E}[x^{(k)}_{i}]$ and $\widehat{Var}[x^{(k)}_{i}]$ reduce the impact of the shift.

In Eq.\,(\ref{eq9}), we present a evaluation method for $\widehat{E}[x^{(k)}_{i}]$. Let $\lambda$ represent the evaluation error, $\lambda = \widehat{E}[x^{(k)}_{i}] \, / \, E[x^{(k)}_{i}]$, where the $\lambda$ is computed on unpruned model.
After pruning, we set

\begin{align}\label{DO-E}
E[x^{(k)}_{i}] = \widehat{E}[x^{(k)}_{i}]\, / \, \lambda 
\end{align}
For $\widehat{Var}[x^{(k)}_{i}]$, it is difficult to evaluate, or the error is large.
We experimentally show that set
\begin{align}\label{DO-V}
Var[x^{(k)}_{i}]=\frac{\widehat{d}_i}{d_i}\times Var[x^{(k)}_{i}],
\end{align}
performs well in most layers, especially in the deeper layers.

\begin{figure}[t]
\centering
\begin{minipage}[!t]{1.\linewidth}
\centering
\subfigure[$l1$-norm-1]{
	\centering
	\includegraphics[scale=0.13]{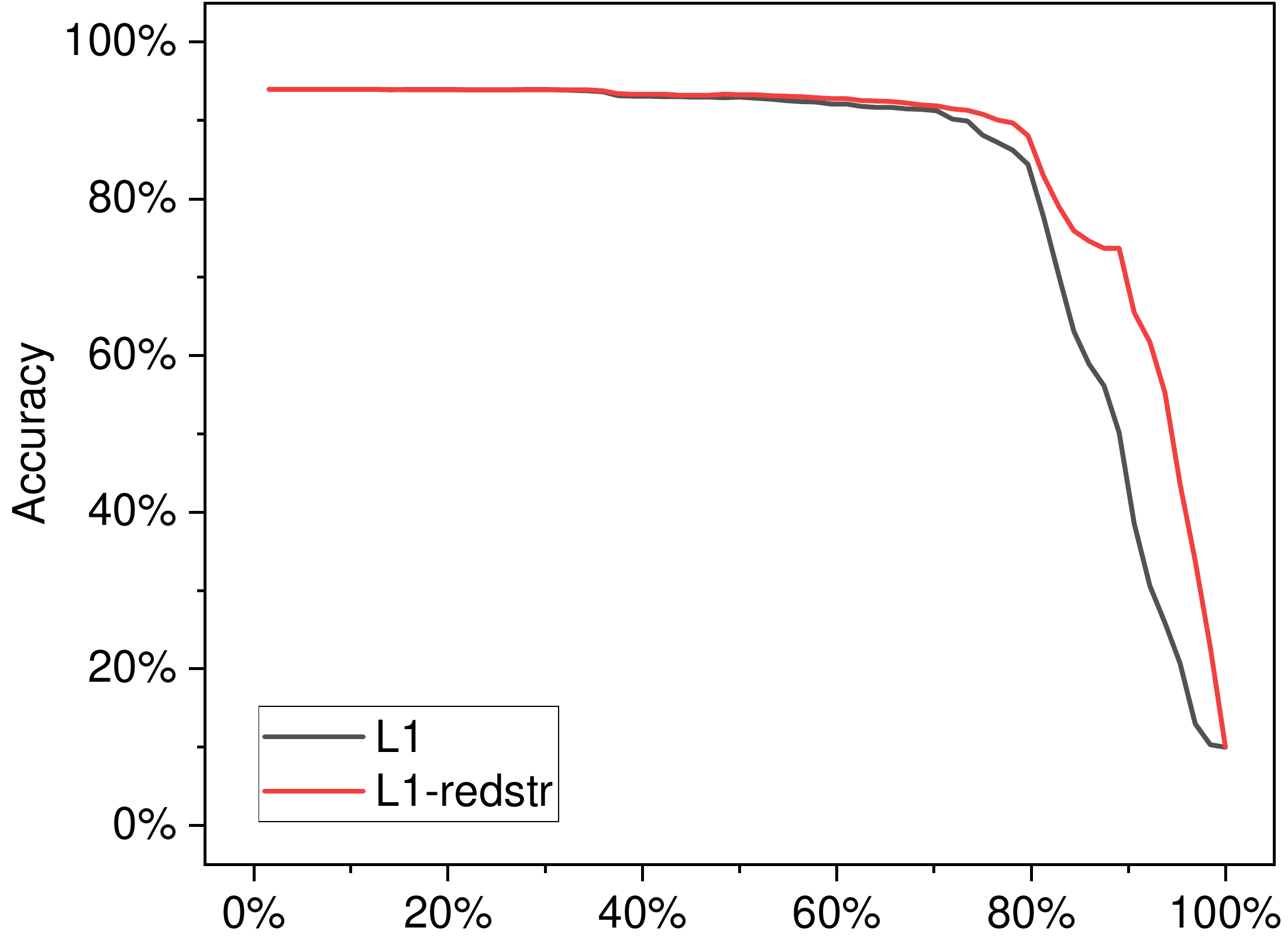}}
\subfigure[FSM-1]{
	\centering
	\includegraphics[scale=0.13]{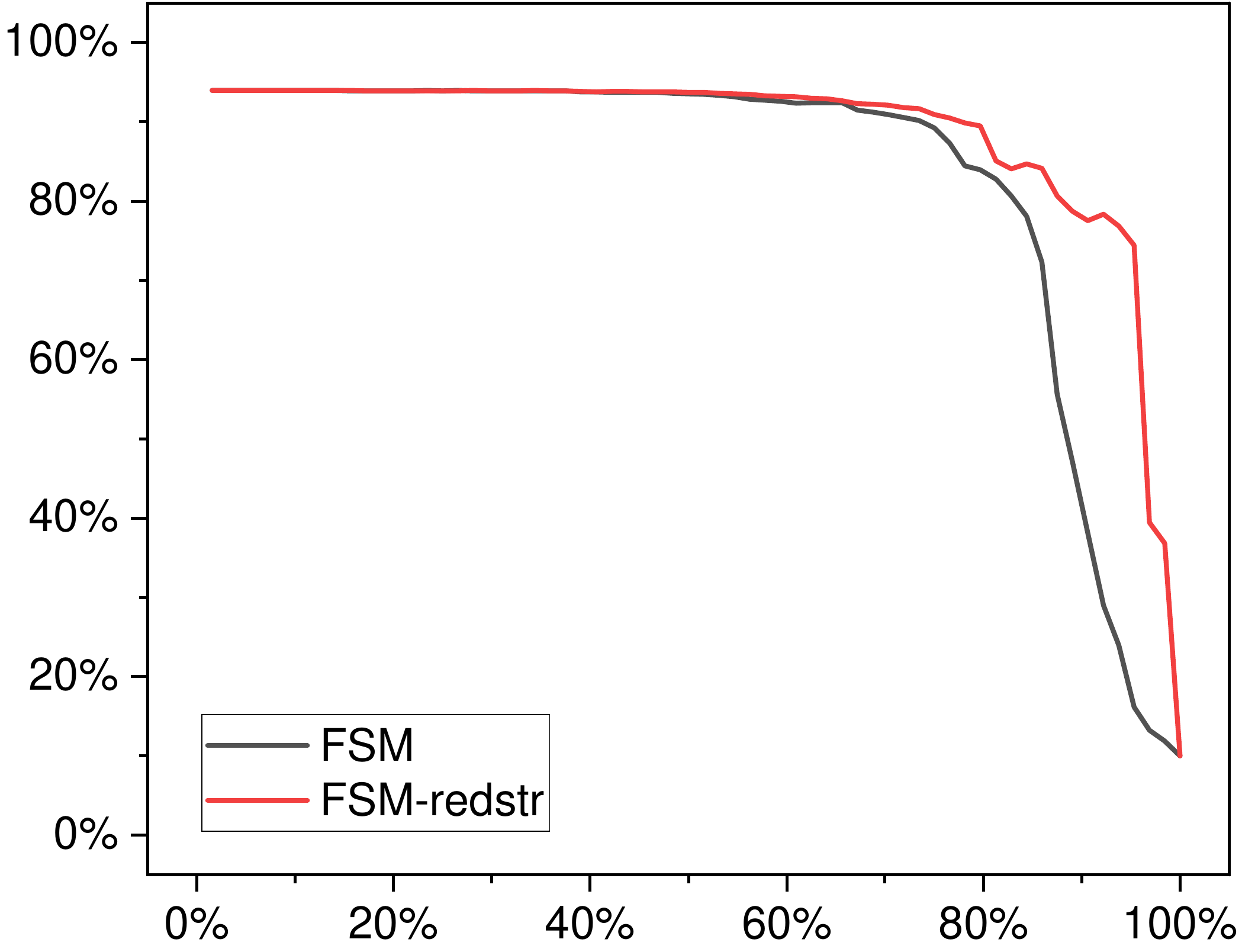}}
\subfigure[HRank-1]{
	\centering
	\includegraphics[scale=0.13]{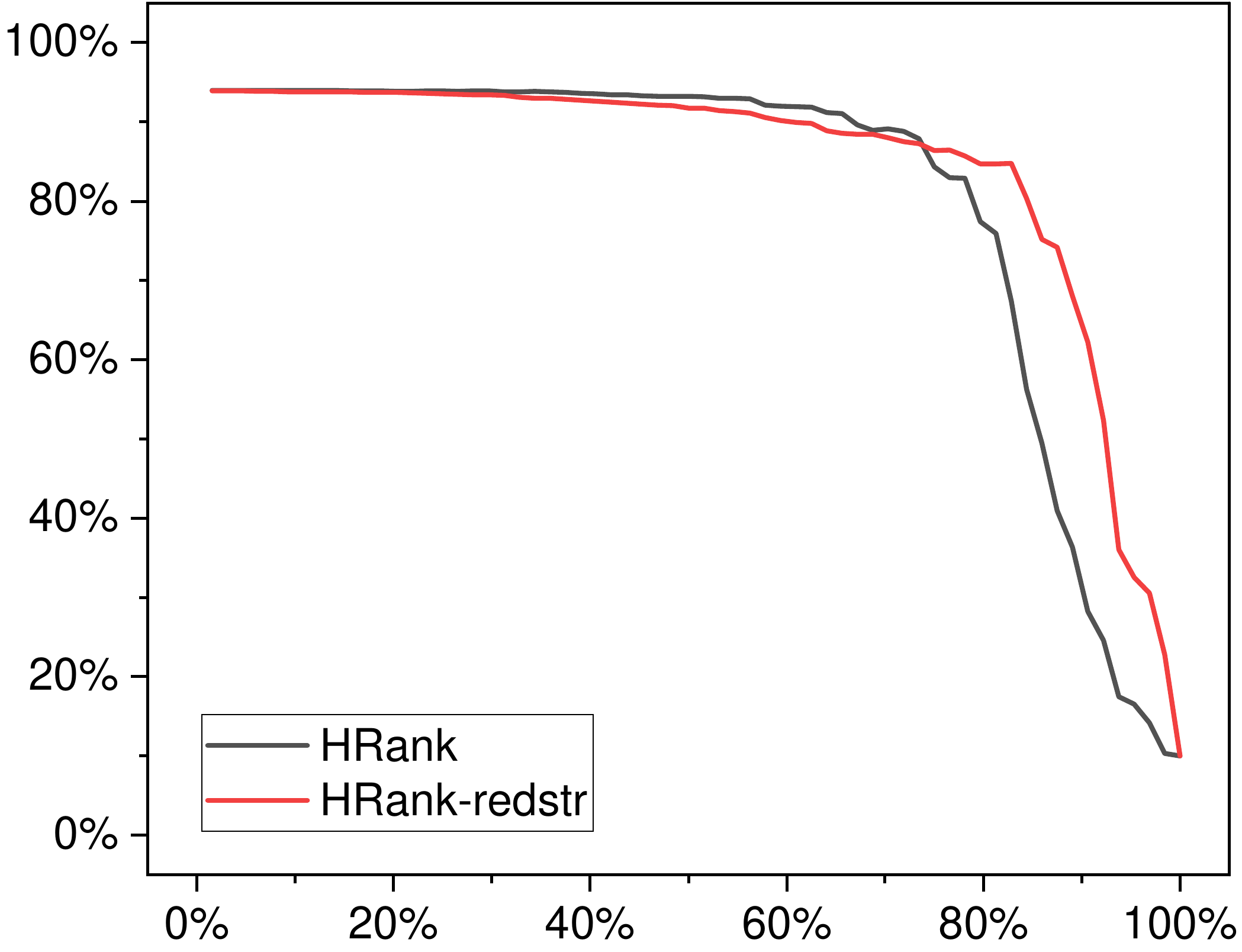}}
\subfigure[FSM-6]{
	\centering
	\includegraphics[scale=0.13]{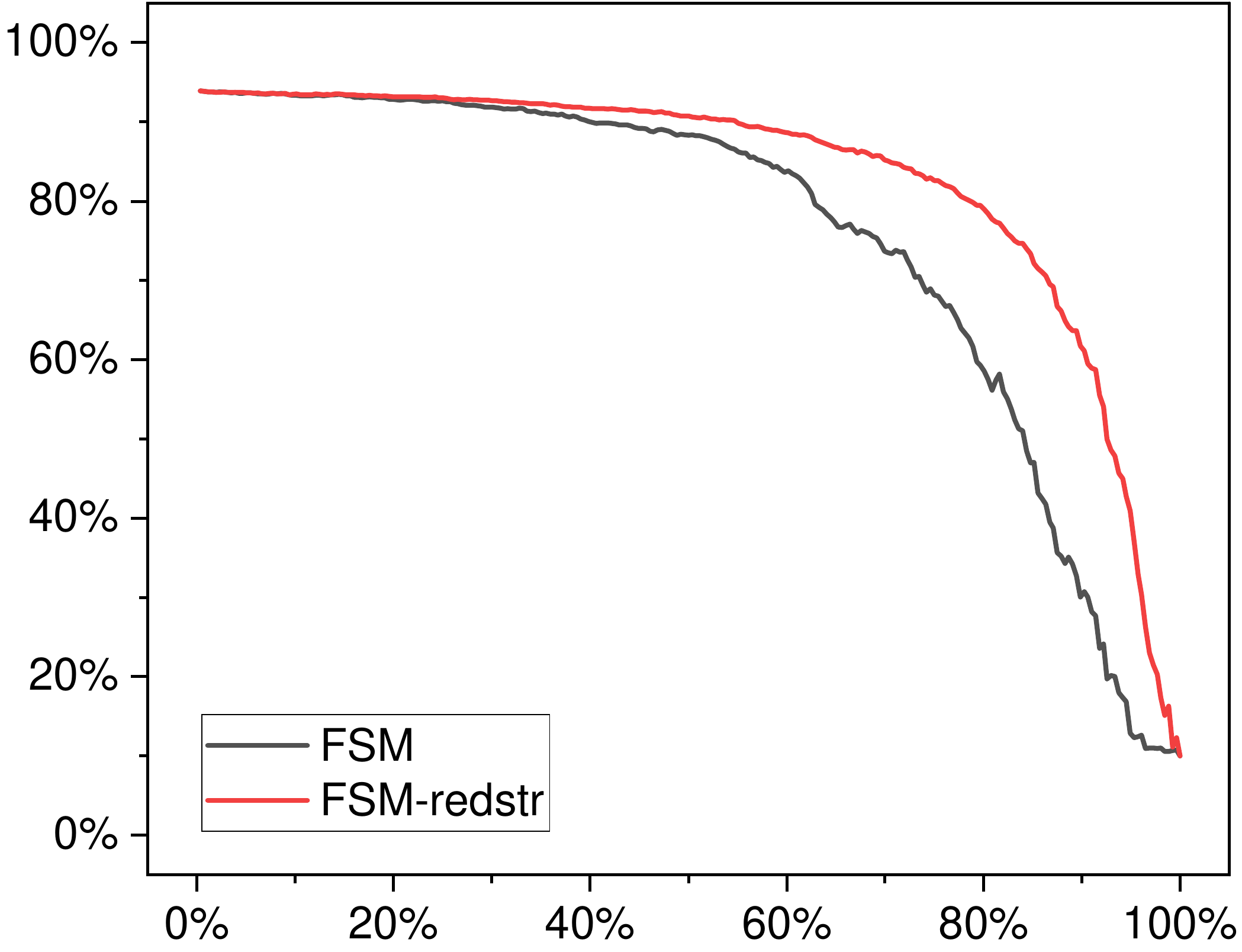}}
\end{minipage}
\begin{minipage}[!t]{1.0\linewidth}
\centering
\subfigure[$l1$-norm-11]{
	\centering
	\includegraphics[scale=0.13]{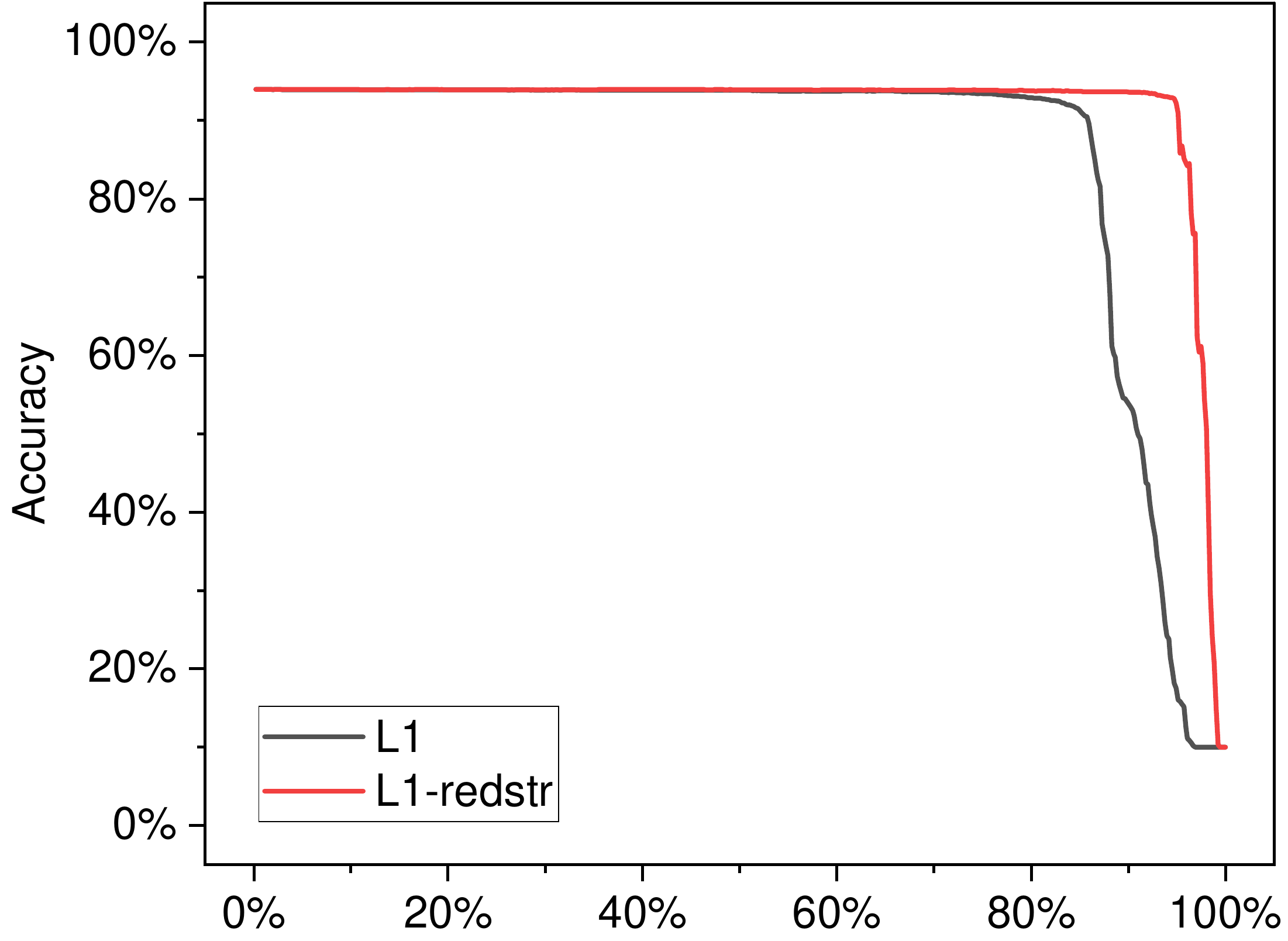}}
\subfigure[HRank-6]{
	\centering
	\includegraphics[scale=0.13]{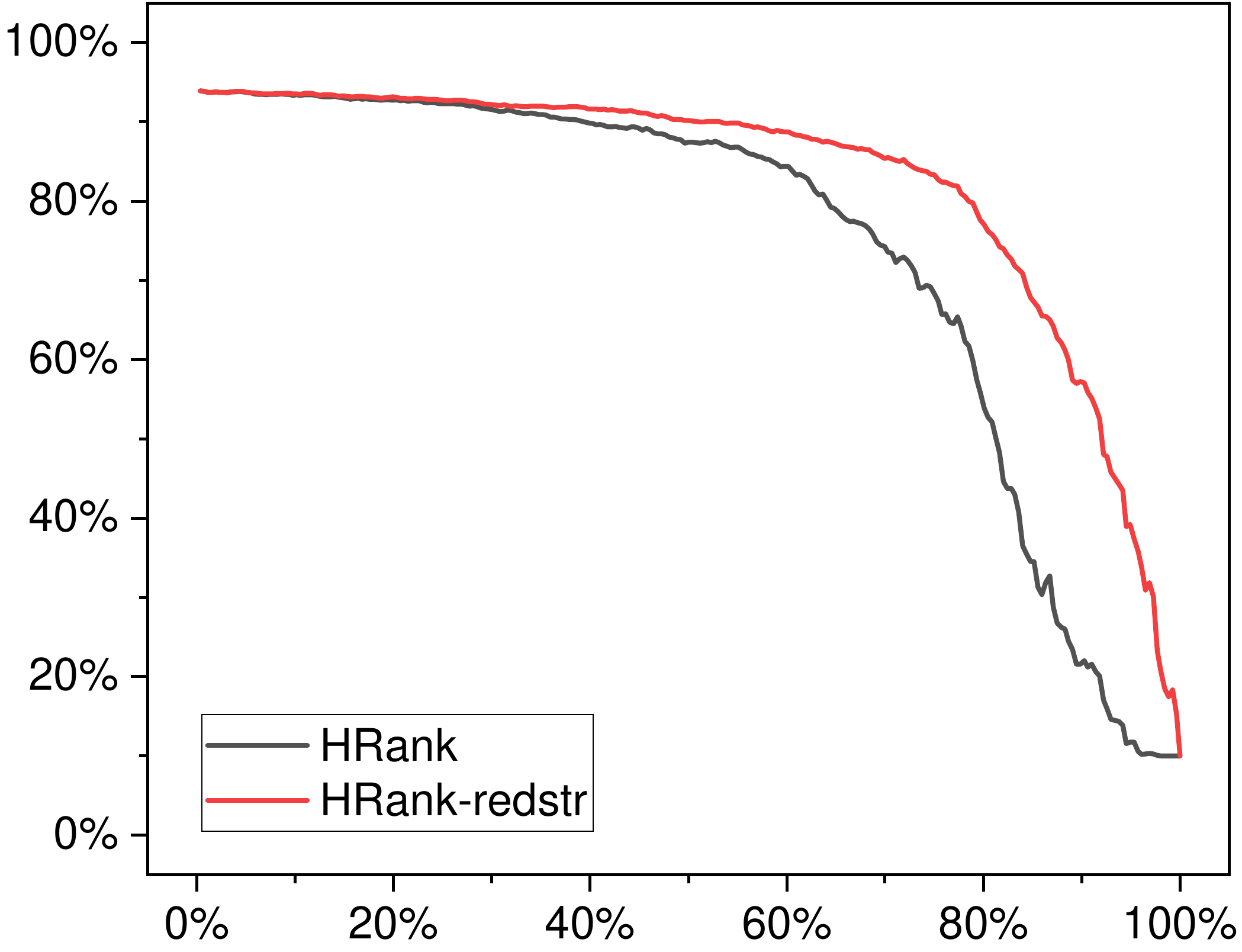}}
\subfigure[FSM-11]{
	\centering
	\includegraphics[scale=0.13]{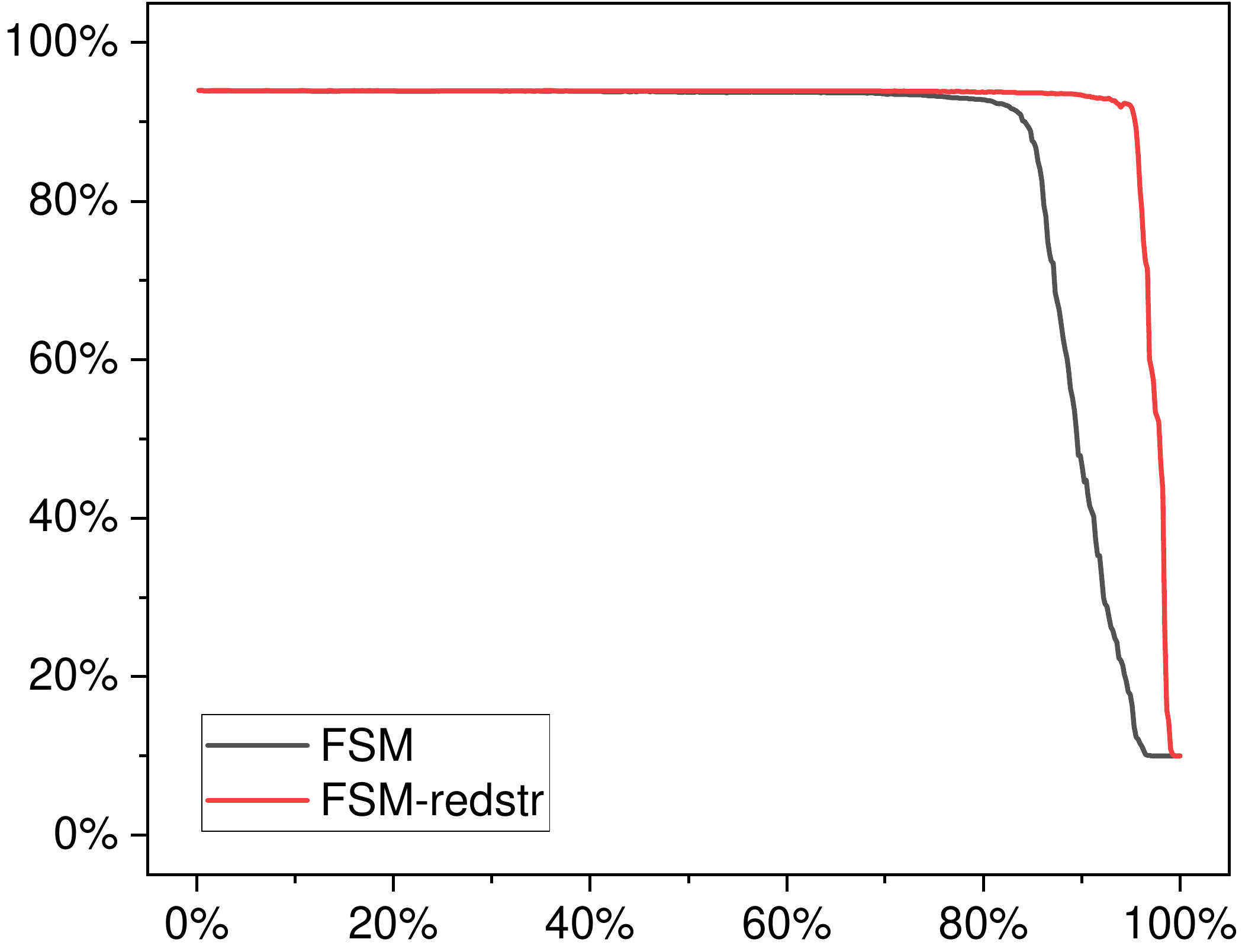}}
\subfigure[HRank-11]{
	\centering
	\includegraphics[scale=0.13]{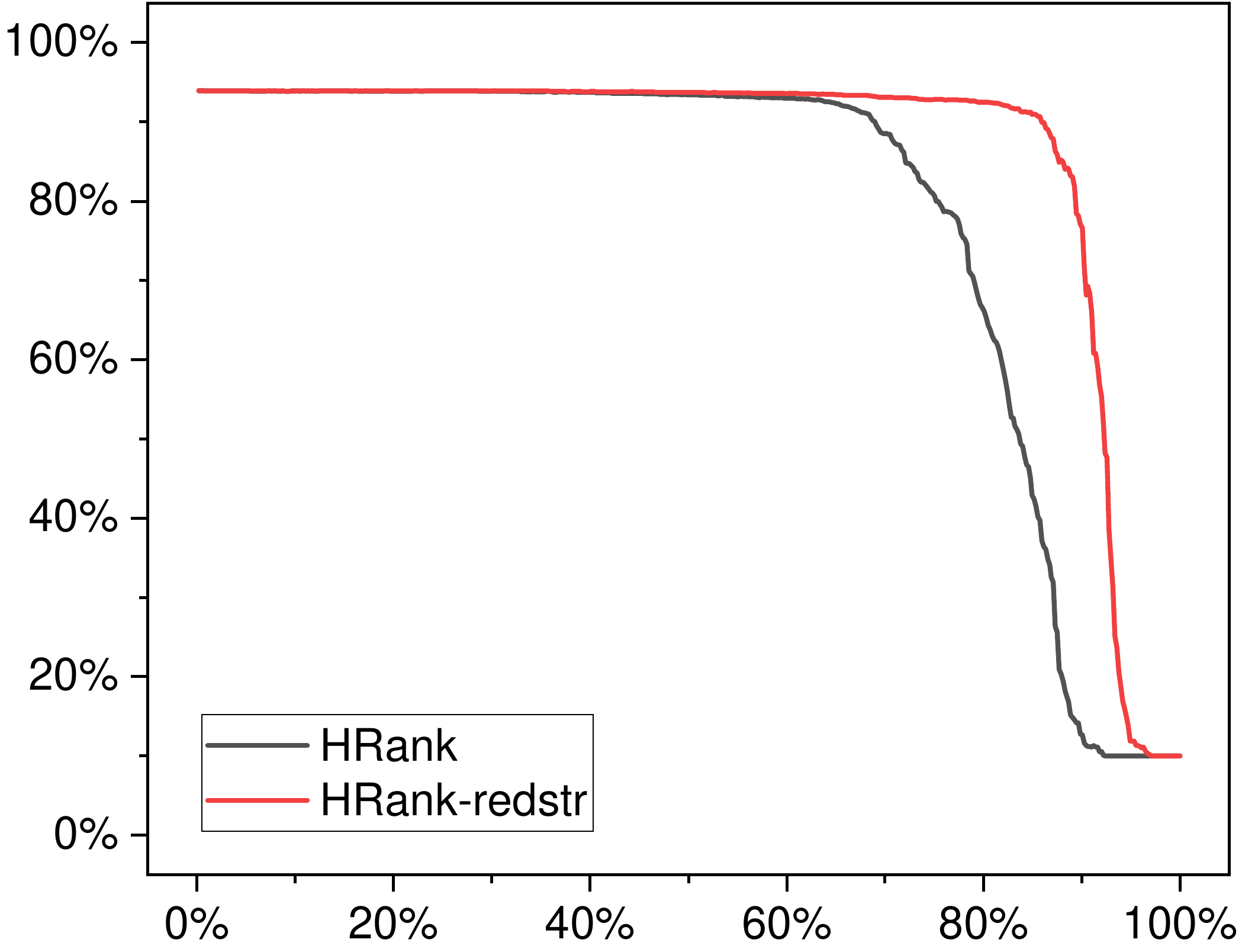}}
\end{minipage}
\caption{Comparison of the accuracy before and after using distribution optimization on VGGNet. Similarly, the horizontal axis represents the compression rate, and the vertical axis is the accuracy. Names such as "HRank-1" mean to adopt the HRank method on the 1st layer of VGGNet. The results suggest that the feature shift is a critical factor that damages model accuracy.}
\label{fig:distribution-optimization}
\end{figure}

One possible explanation is that the differences between high-dimensional features are not obvious, and the statistical properties are approximate.
Random pruning in deep layers produces great results, as shown in Fig.\;\ref{fig:analysis}, supporting this explanation.
Furthermore, as shown in Fig.\;\ref{fig:distribution-optimization}, extensive experiments on different architectures reveal that the proposed distribution optimization strategy is effective and proves that feature shift occurs during network pruning.
Obviously, the algorithm is plug-and-play and may also be combined with other pruning methods.

\subsection{Pruning Procedure}\label{Pruning Procedure}
The pruning procedure is summarized as follows:

\begin{itemize}
  \item[1)] For each channel $x^{(k)}_{i}$ of one layer, we first calculate its output expectation $E[y^{(k)}_{i}]$.
  \item[2)] Then, we calculate the feature shift $\delta(x^{(k)}_{i})$ caused by channel $x^{(k)}_{i}$, according to Eq.\,(\ref{eq15}).
  \item[3)] Sort all channels through $\delta(\cdot)$ and discard those channels with small values.
  \item[4)] After pruning, using Eq.\,(\ref{DO-E}) and Eq.\,(\ref{DO-V}) to recover accuracy.
  \item[5)] Fine-tuning the model one epoch, and back to the first step to prune the next layer.
\end{itemize} 
After all layers have been pruned, we train the pruned model for some epochs.

%
%

\section{Experiments}\label{Experiments}
In this section, in order to verify the effectiveness of the proposed FSM, we conduct extensive experiments on CIFAR-10 \cite{Krizhevsky2009LearningML} and ImageNet \cite{Krizhevsky2012ImageNetCW}. 
Prevalent models, such as ResNet\cite{He2016DeepRL}, GoogLeNet\cite{Szegedy2015GoingDW}, MobileNet-V2\cite{Sandler2018MobileNetV2IR}, and VGGNet\cite{Simonyan2015VeryDC}, are adopted to test the performance.
All experiments are running on Pytorch 1.8 \cite{Paszke2017AutomaticDI} under Intel i7-8700K CPU @3.70GHz and NVIDIA GTX 1080Ti GPU.

\subsection{Implementation Details}\label{Implementation Details}
For all models, the pruning process is performed in two steps:1) pruning the filters layer by layer and fine-tune $1$ epoch for each layer. 2) After pruning, we train the model for some epochs using the stochastic gradient descent algorithm (SGD) with momentum 0.9.
For VGGNet and GoogLeNet on CIFAR-10, we train the model for 200 epochs, in which the initial learning rate, batch size, and weight decay are set to 0.01, 128, and 0, respectively. 
The learning rate is divided by 10 at epochs 100 and 150.
For ResNet-56 on CIFAR-10, we train 300 epochs and set the weight decay to 0.0005.
The learning rate is divided by 10 at epochs 150 and 225.
For ResNet-50, we train the model for 120 epochs with an initial learning rate of 0.01, and the learning rate is divided by 10 at epochs 30, 60, and 90. The batch size is set to 64.
For MobileNet-V2, we fine-tune the pruned model for 150 epochs, and the learning rate is decayed by the cosine annealing scheduler with an initial learning rate of 0.01. The weight decay is set to $4\times 10^{-5}$, following the original MobileNet-V2 paper setup.
We train all the models three times and report the mean.

We set the compression rate of each layer according to its accuracy drop curve, which maintains the model accuracy to the greatest extent possible.
The details are provided in the supplementary.

\setlength{\tabcolsep}{4pt}
\begin{table}[t]
\begin{center}
\caption{Comparison with other prevalent pruning methods on CIFAR-10}
\label{table:CIFAR-10}
\begin{tabular}{c|c|c|c|c|c}
\hline\noalign{\smallskip}
Model & Method & Baseline Acc. & Top-1 Acc. & FLOPs $\downarrow$ & Param. $\downarrow$ \\
\hline
\multirow{7}*{ResNet-56}  & SFP\cite{He2018SoftFP}              & 93.59\% & 93.35\% & 52.6\% & -     \\
                          & CCP\cite{Peng2019CollaborativeCP}   & 93.50\% & 93.42\% & 52.6\% & -     \\
                          & HRank\cite{Lin2020HRankFP}          & 93.26\% & 92.17\% & 50.0\% & 42.4\%\\
                          & NPPM\cite{Gao2021NetworkPV}         & 93.04\% & 93.40\% & 50.0\% & -      \\  
                          & DHP\cite{Li2020DHPDM}               & -       & 93.58\% & 49.0\% & 41.6\%\\
                          & SCP\cite{Kang2020OperationAwareSC}  & 93.69   & 93.23\% & 51.5\% & -\\
                          & \textbf{FSM}(ours)                  & \textbf{93.26}\% & \textbf{93.63}\% & \textbf{51.2}\% & \textbf{43.6}\% \\  
                          & \textbf{FSM}(ours)                  & \textbf{93.26}\% & \textbf{92.76}\% & \textbf{68.2}\% & \textbf{68.5}\% \\
\hline
\multirow{5}*{VGGNet}     & L1\cite{Li2017PruningFF}            & 93.25\% & 93.40\% & 34.2\% & 63.3\%\\
                          & GAL\cite{Lin2019TowardsOS}          & 93.96\% & 92.03\% & 39.6\% & 77.2\%\\
                          & HRank\cite{Lin2020HRankFP}          & 93.96\% & 93.42\% & 53.7\% & 82.9\%\\
                          & EEMC\cite{Zhang2021ExplorationAE}   & 93.36\% & 93.63\% & 56.6\% & - \\

                          & \textbf{FSM}(ours)                  & \textbf{93.96}\% & \textbf{93.73}\% & \textbf{66.0}\% & \textbf{86.3}\%\\
                          & \textbf{FSM}(ours)                  & \textbf{93.96}\% & \textbf{92.86}\% & \textbf{81.0}\% & \textbf{90.6}\%\\
\hline
\multirow{5}*{GoogLeNet}  & L1\cite{Li2017PruningFF}            & -       & 94.54\% & 32.9\% & 42.9\%\\
                          & GAL\cite{Lin2019TowardsOS}          & 95.05\% & 93.93\% & 38.2\% & 49.3\%\\
                          & HRank\cite{Lin2020HRankFP}          & 95.05\% & 94.53\% & 54.6\% & 55.4\%\\
                          & \textbf{FSM}(ours)                  & \textbf{95.05}\% & \textbf{94.72}\% & \textbf{62.9}\% & \textbf{55.5}\%\\
                          & \textbf{FSM}(ours)                  & \textbf{95.05}\% & \textbf{94.29}\% & \textbf{75.4}\% & \textbf{64.6}\%\\
\hline
\end{tabular}
\end{center}
\end{table}
\setlength{\tabcolsep}{4pt}
\begin{table}
\begin{center}
\caption{Comparison with other prevalent pruning methods on ImageNet}
\label{table:ImageNet}
\begin{tabular}{c|c|c|c|c|c|c}
\hline\noalign{\smallskip}
Model & Method & Top-1 Acc. & Top-5 Acc. & $\Delta$ Top-1 & $\Delta$ Top-5 & FLOPs $\downarrow$  \\
\hline
 \multirow{10}*{ResNet-50}  &  DCP\cite{Zhuang2018DiscriminationawareCP}    & 74.95\% & 92.32\% & -1.06\% & -0.61\% & 55.6\% \\  
                            &  CCP\cite{Peng2019CollaborativeCP}            & 75.21\% & 92.42\% & -0.94\% & -0.45\% & 54.1\% \\ 
                           &   Meta\cite{Liu2019MetaPruningML}              & 75.40\% & -       & -1.20\% & -       & 51.2\% \\ 
                           &   GBN\cite{You2019GateDG}                      & 75.18\% & 92.41\% & -0.67\% & -0.26\% & 55.1\% \\ 
                           &   BNFI\cite{Oh2022BatchNT}                     & 75.02\% & - & -1.29\% & - & 52.8\% \\ 
                           &   HRank\cite{Lin2020HRankFP}                   & 74.98\% & 92.44\% & -1.17\% & -0.54\% & 43.8\% \\ 
                           &   SCP\cite{Kang2020OperationAwareSC}           & 75.27\% & 92.30\% & -0.62\% & -0.68\% & 54.3\% \\ 
                           &   SRR-GR\cite{Wang2021ConvolutionalNN}         & 75.11\% & 92.35\% & -1.02\% & -0.51\% & 55.1\% \\ 
                           &   GReg\cite{wang2020neural}                    & 75.16\% & - & -0.97\% & - & 56.7\% \\ 
                           &   \textbf{FSM}(ours)                           & \textbf{75.43}\% & \textbf{92.45}\% & \textbf{-0.66}\% & \textbf{-0.53}\% & \textbf{57.2}\% \\         
\hline
\multirow{3}*{MobileNet-V2}&  CC\cite{wang2021convolutional}                & 70.91\% & - & -0.89\% & - & 30.7\% \\  
                           &  BNFI\cite{Oh2022BatchNT}                      & 70.97\% & - & -1.22\% & - & 30.0\% \\
                           &  \textbf{FSM}(ours)                            & \textbf{71.18}\% & \textbf{89.81}\% & \textbf{-0.70}\% & \textbf{-0.48}\% & \textbf{30.6}\% \\ 
\hline
\end{tabular}
\end{center}
\end{table}
\subsection{Results and Analysis}\label{Results and Analysis}
{\bf CIFAR-10.} 
In Table~\ref{table:CIFAR-10}, we compare the proposed FSM method with other prevalent algorithms on VGGNet, GoogLeNet, and ResNet-56.
Our method achieves a signiﬁcantly compression efficiency on reductions of FLOPs and  parameters, but with higher accuracy.
For example, compared with HRank, FSM yields a better accuracy (93.73\% vs. 93.42\%) under a greater FLOPs reduction (66.0\% vs. 53.7\%) and parameters reduction (86.3\% vs. 82.9\%).
For ResNet-56 on CIFAR-10, FSM prunes 51.2\% of FLOPs and 43.6\% of parameters, but the accuracy improved by 0.37\%. 
NPPM and DHP have been proposed recently and they have a great performance in network compression.
Compared with NPPM, our method shows a better performance, and with a higher FLOPs reduction. 
Compared with DHP, under similar accuracy, FSM performs better at the reduction of FLOPs (51.2\% vs. 49.0\%) and parameters (43.6\% vs.  41.6\%).
For GoogLeNet, FSM outperforms HRank and GAL in all aspects. 
With over 62\% FLOPs reduction, FSM still maintains 94.72\% accuracy.

Furthermore, we verified the performance of FSM at high compression rates.
For instance, with more than 90\% of the parameters discarded, FSM reduces the accuracy by only 1.1\% on VGGNet.
The same results can be observed on ResNet-56 and GoogLeNet.

{\bf ImageNet 2012.} 
ImageNet 2012 has over 1.28 million training images and 50,000 validation images divided into 1,000 categories.
ResNet-50, compared to VGGNet and ResNet-56, has a larger feature size. 
To verify the applicability of the proposed FSM, on ResNet-50, we test the performance at different compression rates and different layer-depth, by comparing it with L1 and HRank, as shown in Fig.\;\ref{fig:analysis}.
The results show that FSM successfully picks out the more important channels, even for large-size features. In contrast, L1 and HRank performed poorly, even less well than the random method.
In most cases, the proposed FSM shows the best performance.

In Table~\ref{table:ImageNet}, we compare the proposed FSM method with other prevalent methods on ResNet-50.
Our method achieves significant performance, reducing the FLOPs by more than 57\% with only 0.66\% loss in Top-1 accuracy.
Compared with HRank, our method shows a better performance (75.43\% vs. 74.98\%), and with a higher FLOPs reduction (57.2\% vs. 43.8\%). 
Compared with GBN and DCP, under similar FLOPs reduction of about 55\%, FSM obtains better accuracy.
Our method reduces FLOPs by 6\% more than MetaPruning under similar Top-1 accuracy.
More comparison details can be found in Table~\ref{table:ImageNet}.

MobileNet-V2 is a memory-efficient model that is suitable for mobile devices. 
However, to further compress it is challenging while maintaining the model accuracy.
When pruning around 30\% of FLOPs, FSM achieves better performance (71.18\%) than CC (70.91\%) and BNFI (70.97\%) on Top-1 accuracy.
The results demonstrate that FSM has excellent performance on the large-scale ImageNet dataset and shows higher applicability.

\setlength{\tabcolsep}{4pt}
\begin{table}[t]
\begin{center}
\caption{The practical speedup for the pruned model over the unpruned model. "T": inference time. "S": practical speedup on GPU (NVIDIA 1080Ti).}
\label{table:speedup}
\begin{tabular}{c|cccc}
\hline\noalign{\smallskip}
Model &   FLOPs(\%) $\downarrow$ &  Param.(\%) $\downarrow$ & T(ms) & S($\times$)   \\
\hline
 \multirow{2}*{GoogLeNet} & 62.9 & 55.5 & 5.03($\pm 0.01$)& 1.47$\times$\\ 
     & 75.4 & 64.7 & 4.58($\pm 0.02$)& 1.62$\times$\\  
\hline
 ResNet-50 & 57.2 & 42.8 & 13.81($\pm 0.04$)& 1.30$\times$\\
\hline
\end{tabular}
\end{center}
\end{table}
{\bf The Speedup on GPU.} 
To show the practical speedup of our method in real scenarios, we measure the inference time by time/images on the NVIDIA 1080Ti GPU.
As shown in Table~\ref{table:speedup}, the proposed FSM achieves 1.62$\times$ and 1.30$\times$ speedup with batch size 64 on GoogLeNet and ResNet-50, respectively.

%
%

\section{Ablation Study}\label{Ablation Study}
\subsection{Effect of the Activation Function}\label{Effect of the Activation Function}
\begin{figure}[t]
\centering
\begin{minipage}[!t]{1.\linewidth}
\centering
\subfigure[layer-1]{
	\centering
	\includegraphics[scale=0.17]{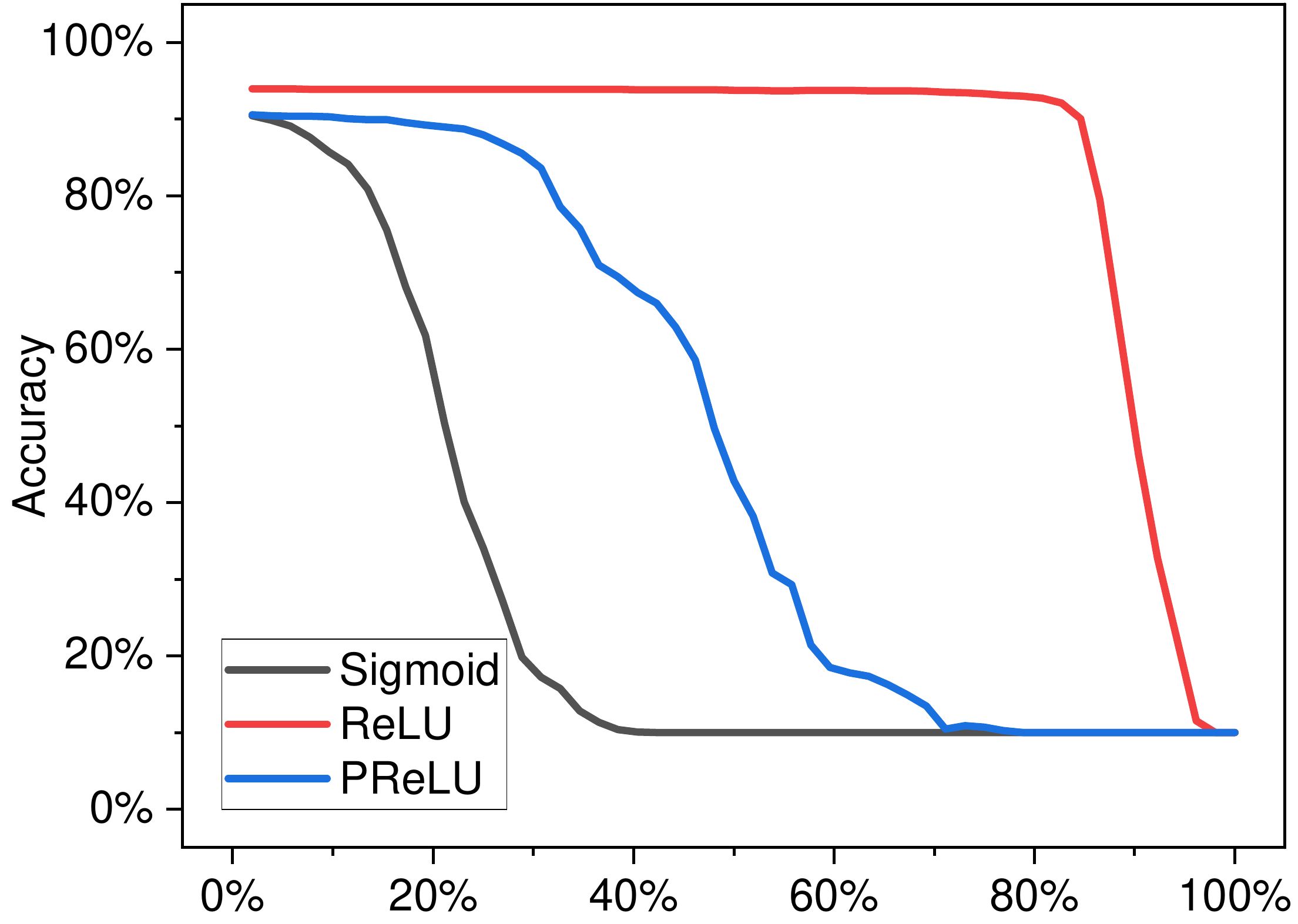}}
\subfigure[layer-6]{
	\centering
	\includegraphics[scale=0.17]{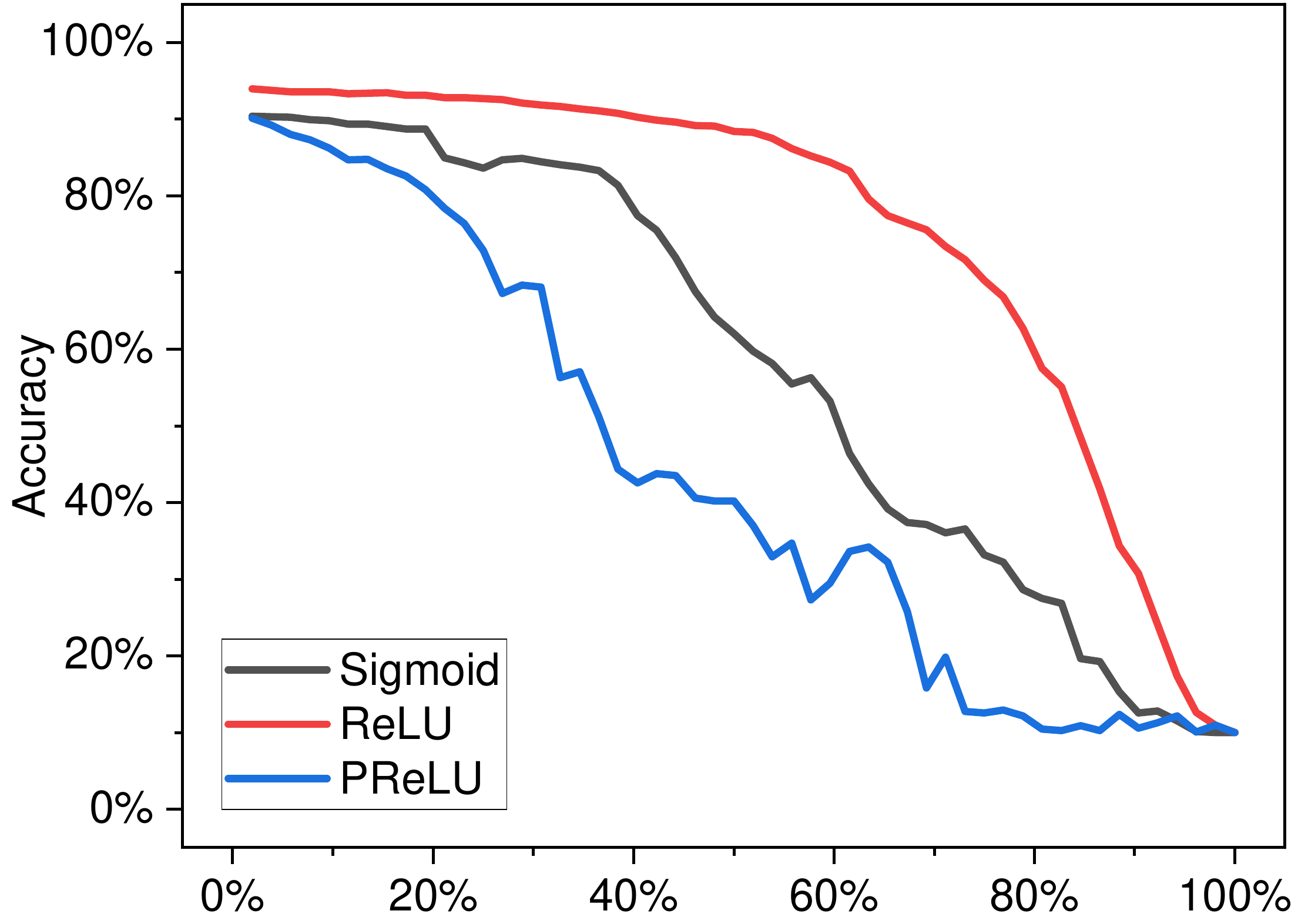}}
\subfigure[layer-11]{
	\centering
	\includegraphics[scale=0.17]{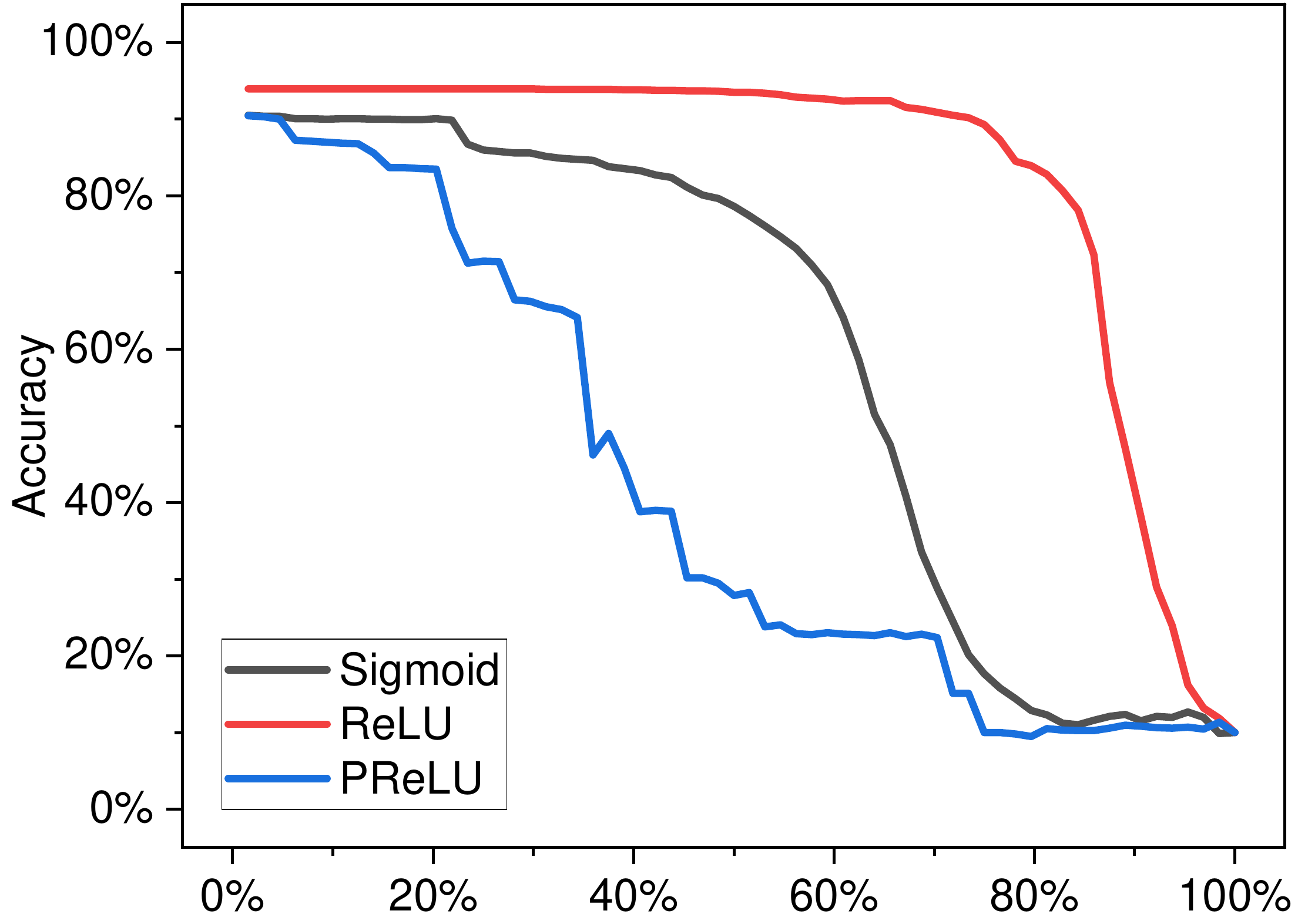}}
\end{minipage}
\caption{The effect of different activation functions on channel pruning for VGG. For each subﬁgure, the horizontal axis is the compression rate, and the vertical axis is the accuracy.}
\label{fig:ablation-1}
\end{figure}

In Section\;\ref{Filter Pruning Analysis}, we illustrate that the activation function affects the distribution of the features. 
In channel pruning, the threshold property of the ReLU activation function results in changes in the feature activation state.
In this section, we experiment with the performance of three popular activation functions (Sigmoid, ReLU, and PReLU \cite{He2015DelvingDI}) on different layers in VGGNet.
As shown in Fig.\;\ref{fig:ablation-1}, the trend of accuracy with Sigmoid and PReLU changes smoother than ReLU.
Although they do not have the problem of gradient plunges, there is a large loss of accuracy at small compression rates.
In contrast, ReLU achieves better performance at most rates.
In particular, $PReLU(x)=max(0,x)+a\times min(0,x)$
, and $Sigmoid(x)=\sigma(x)=\frac{1}{1+e^{-x}}$.
Different from the ReLU, their gradient is greater than $0$ when $x < 0$. 
Some values should reach 0 under ReLU, but return a negative value under PReLU, which leads to more changes in features.
This explains why PReLU loses more accuracy when the feature shift occurs compared to Sigmoid and ReLU.
In general, models with ReLU activation functions are more tolerable for channel pruning than Sigmoid and PReLU.
\subsection{Error of the Feature Shift Evaluation}\label{Error of the Feature Shift Evaluation}
\begin{figure}[t]
\centering
\begin{minipage}[!t]{1.\linewidth}
\centering
\subfigure[VGGNet-1]{
	\centering
	\includegraphics[scale=0.13]{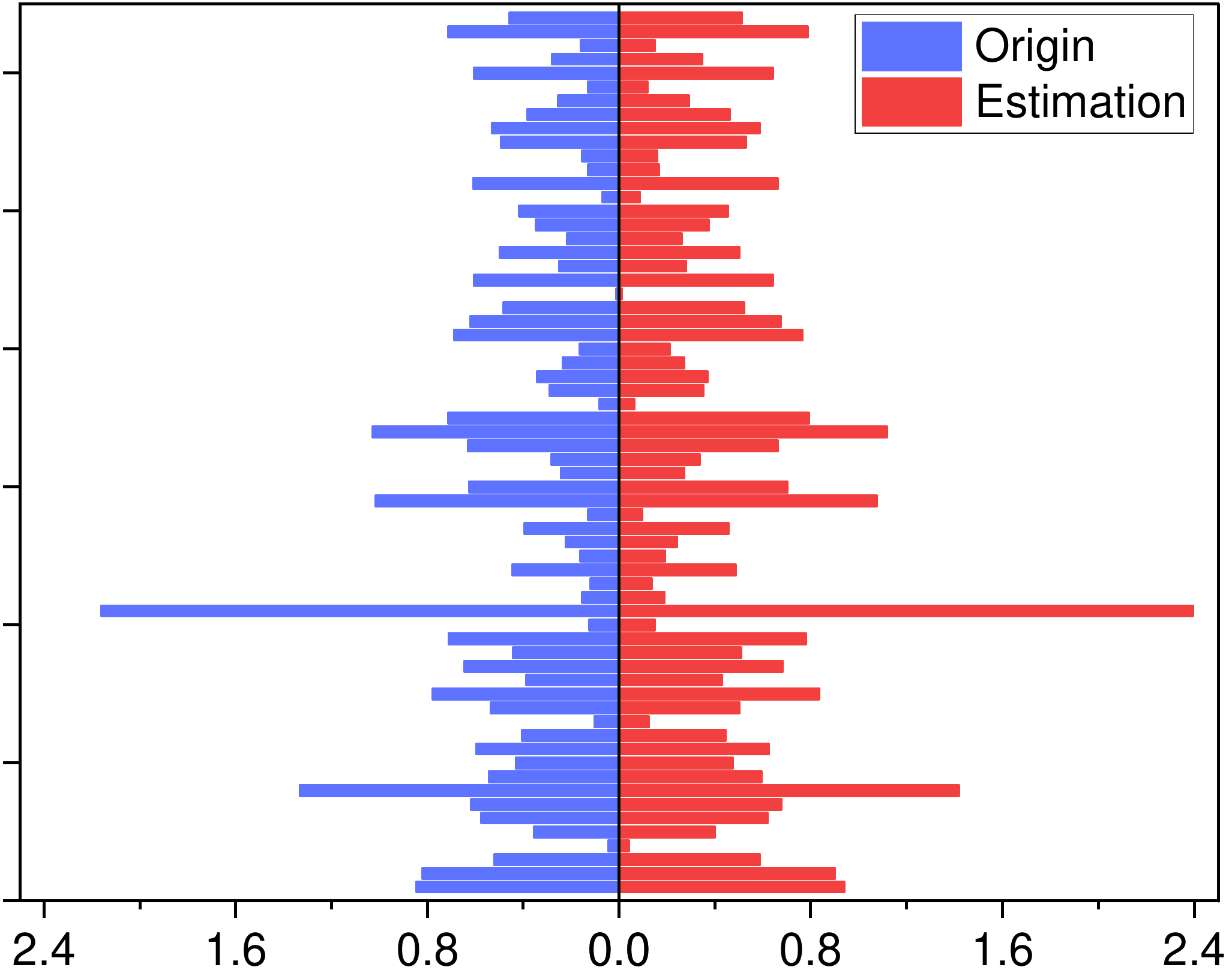}}
\subfigure[VGGNet-3]{
	\centering
	\includegraphics[scale=0.13]{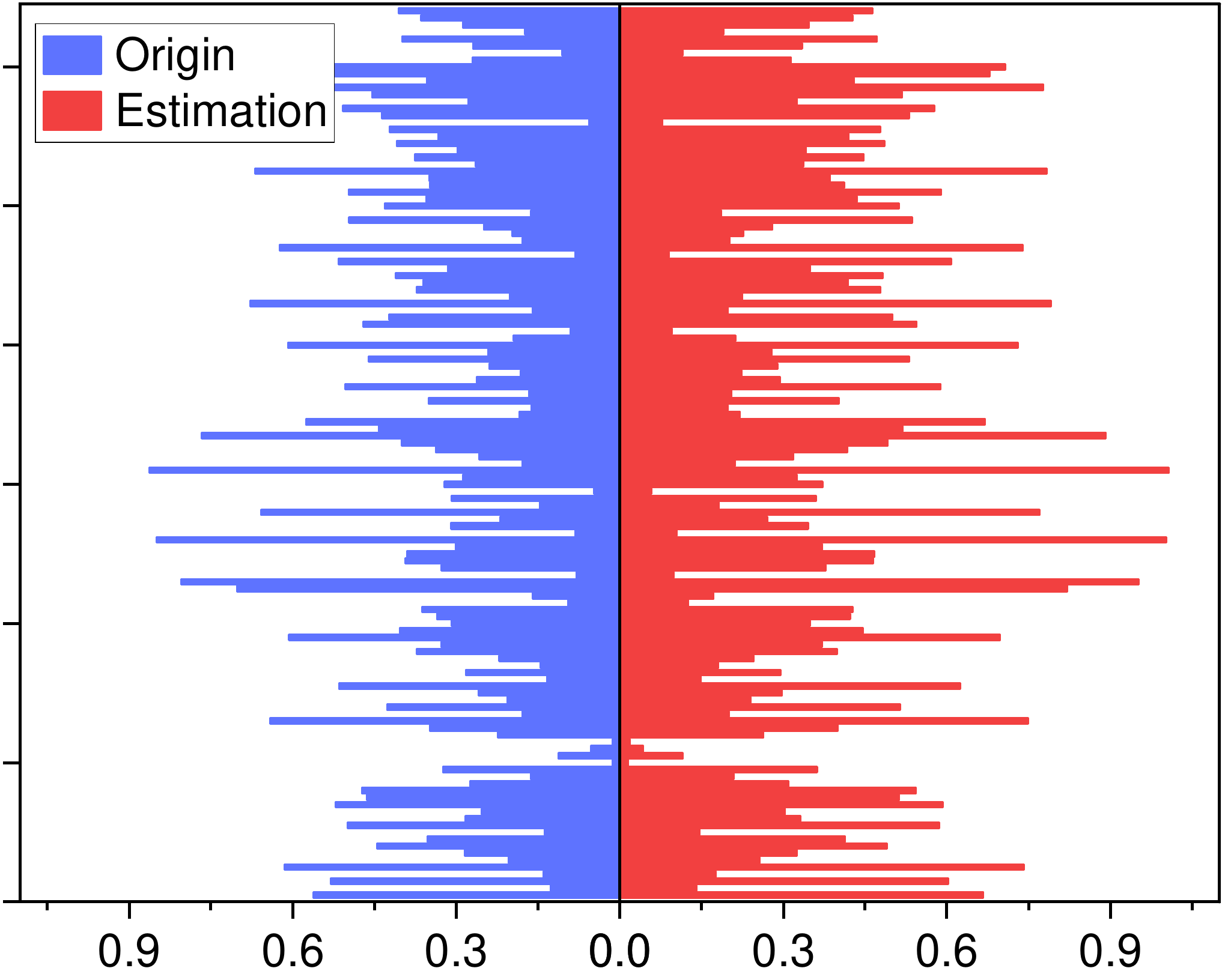}}
\subfigure[VGGNet-6]{
	\centering
	\includegraphics[scale=0.13]{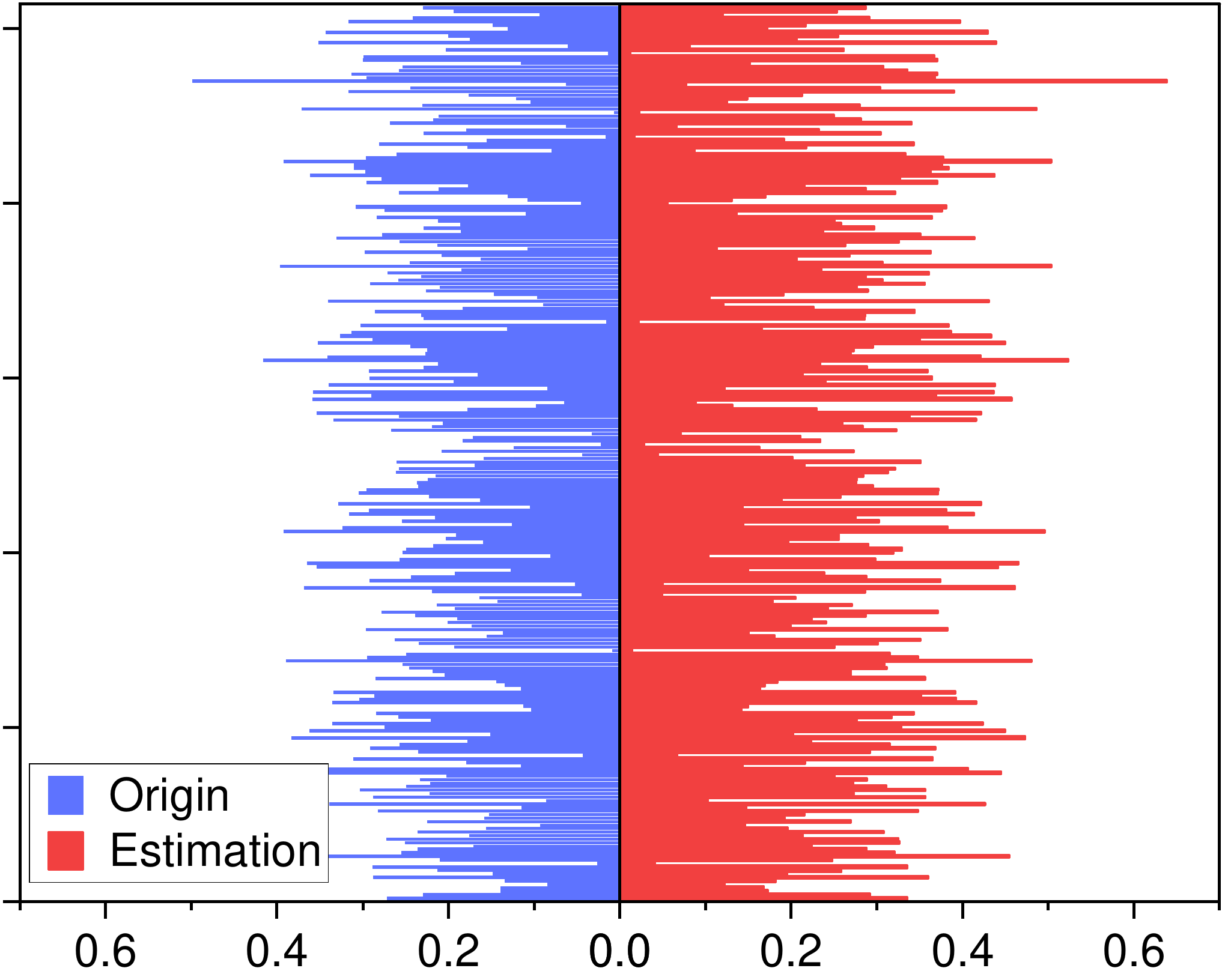}}
\subfigure[ResNet-50-3]{
	\centering
	\includegraphics[scale=0.13]{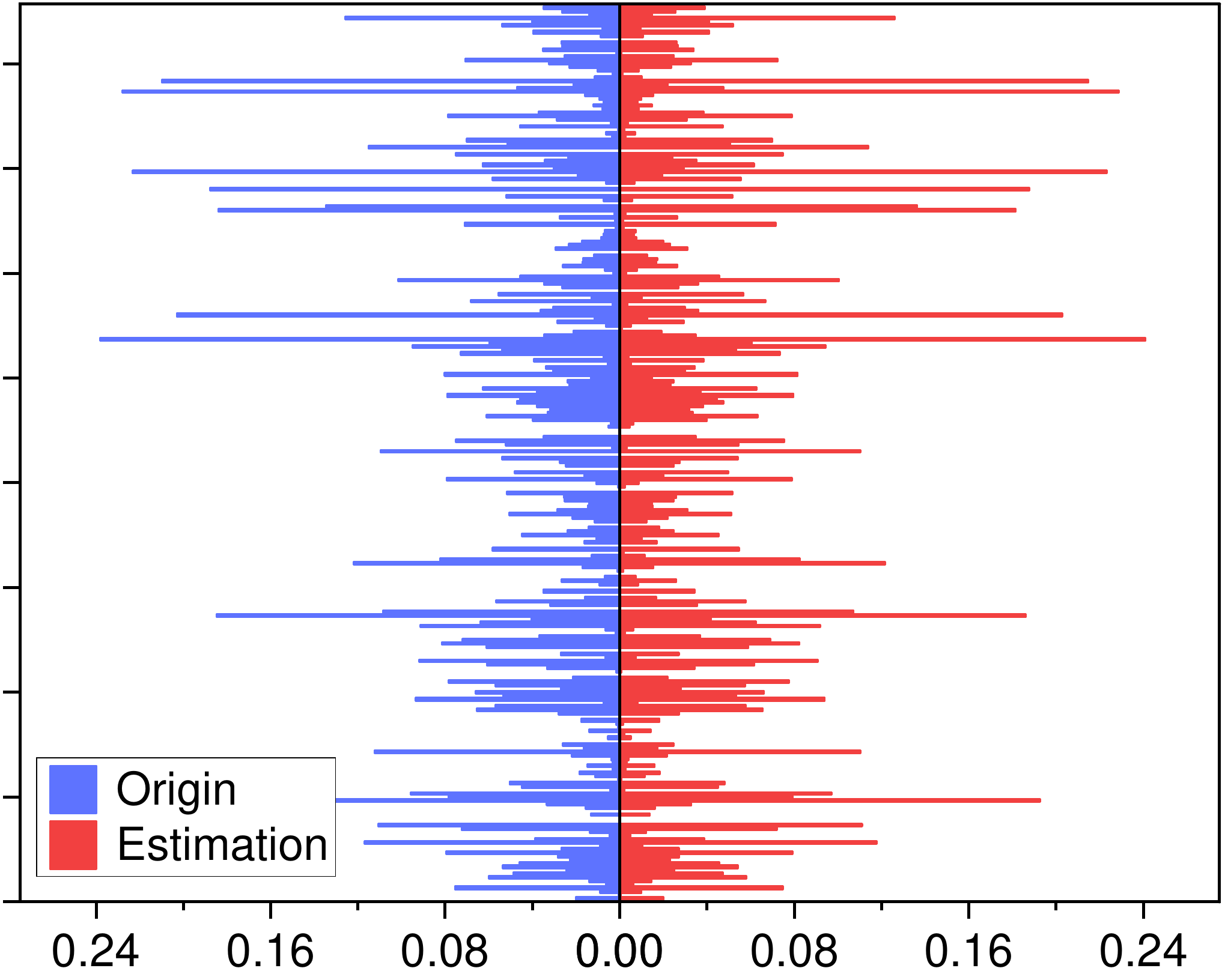}}
\end{minipage}
\begin{minipage}[!t]{1.0\linewidth}
\centering
\subfigure[ResNet-50-7]{
	\centering
	\includegraphics[scale=0.13]{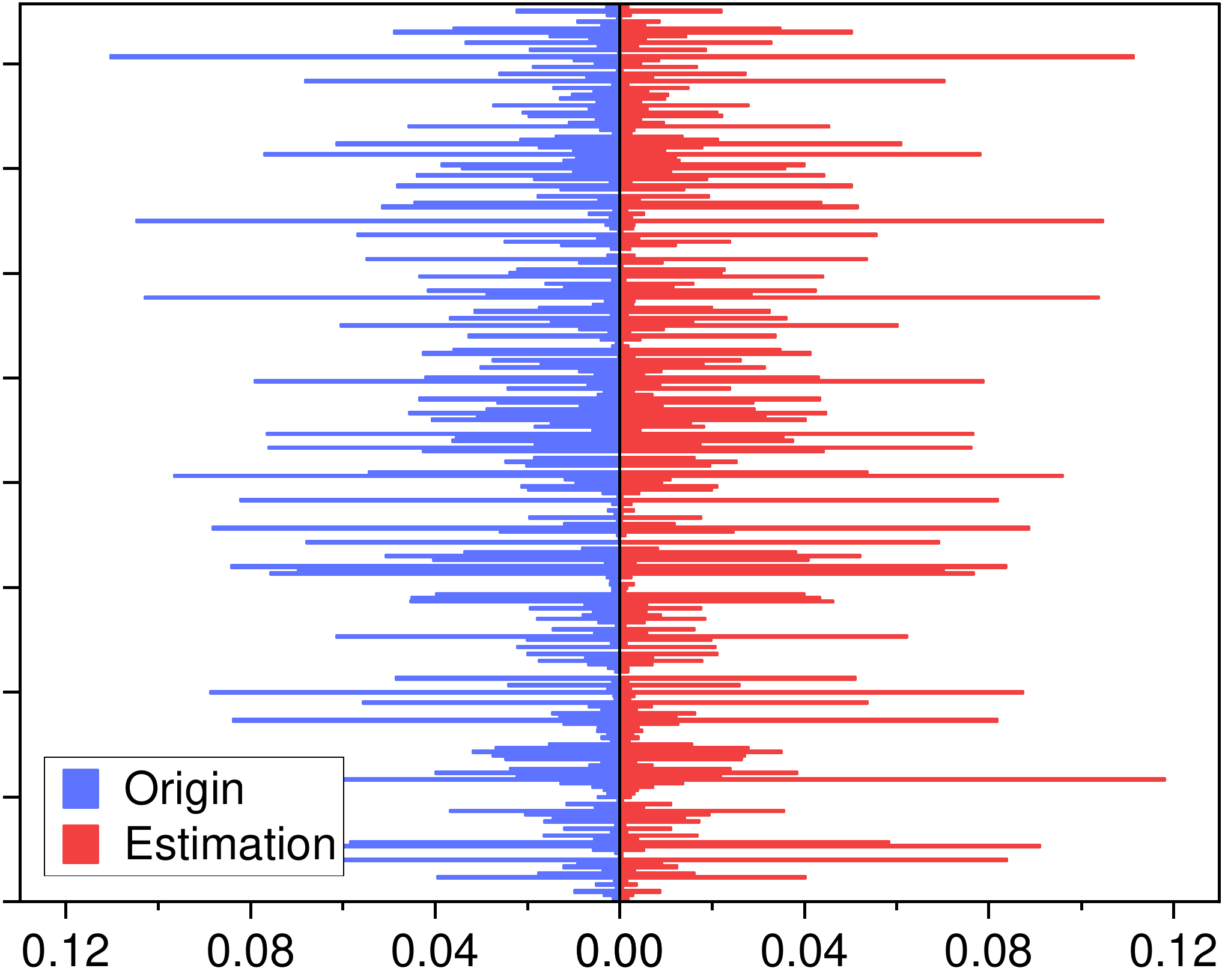}}
\subfigure[GoogLeNet-3]{
	\centering
	\includegraphics[scale=0.13]{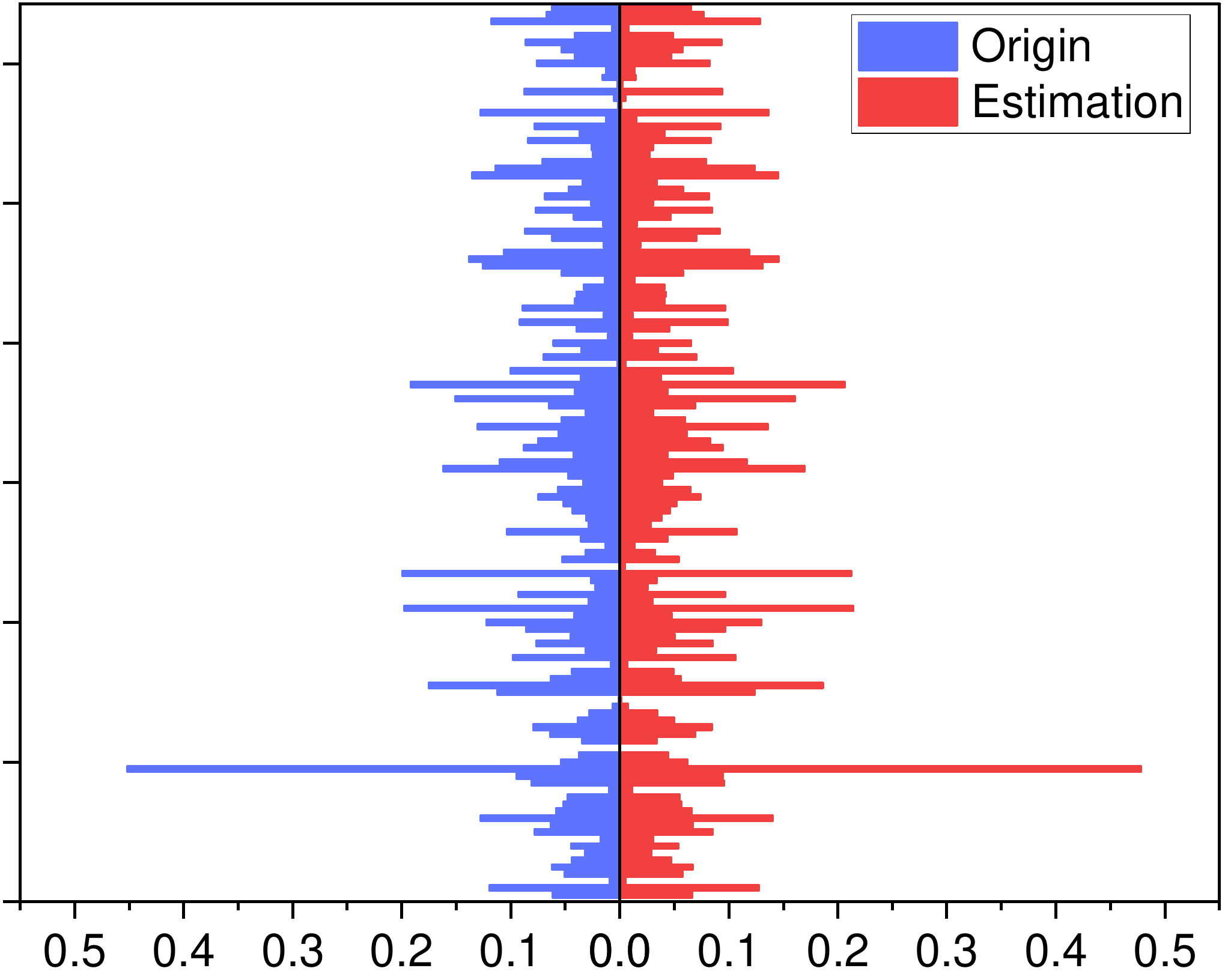}}
\subfigure[GoogLeNet-5]{
	\centering
	\includegraphics[scale=0.13]{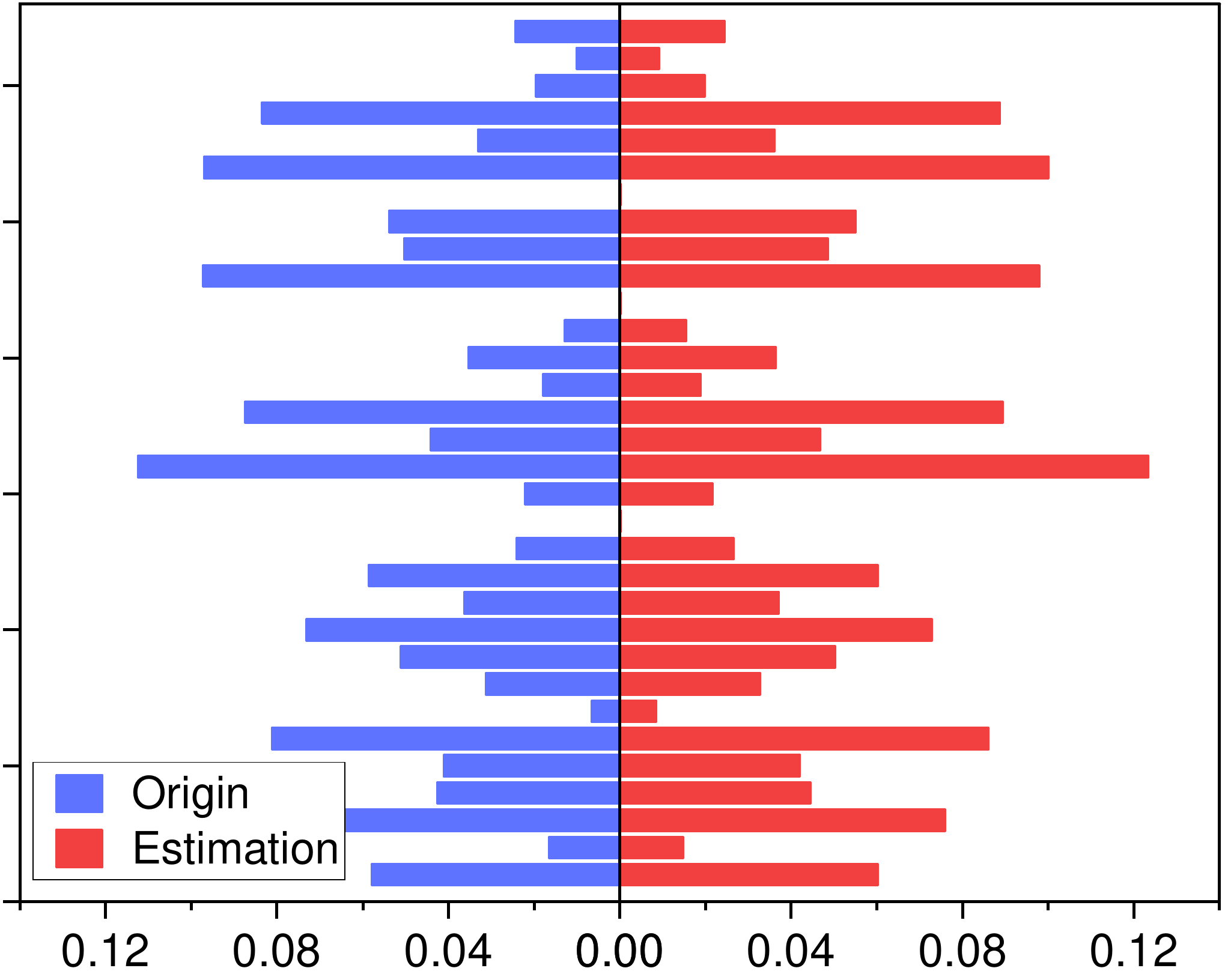}}
\subfigure[GoogLeNet-6]{
	\centering
	\includegraphics[scale=0.13]{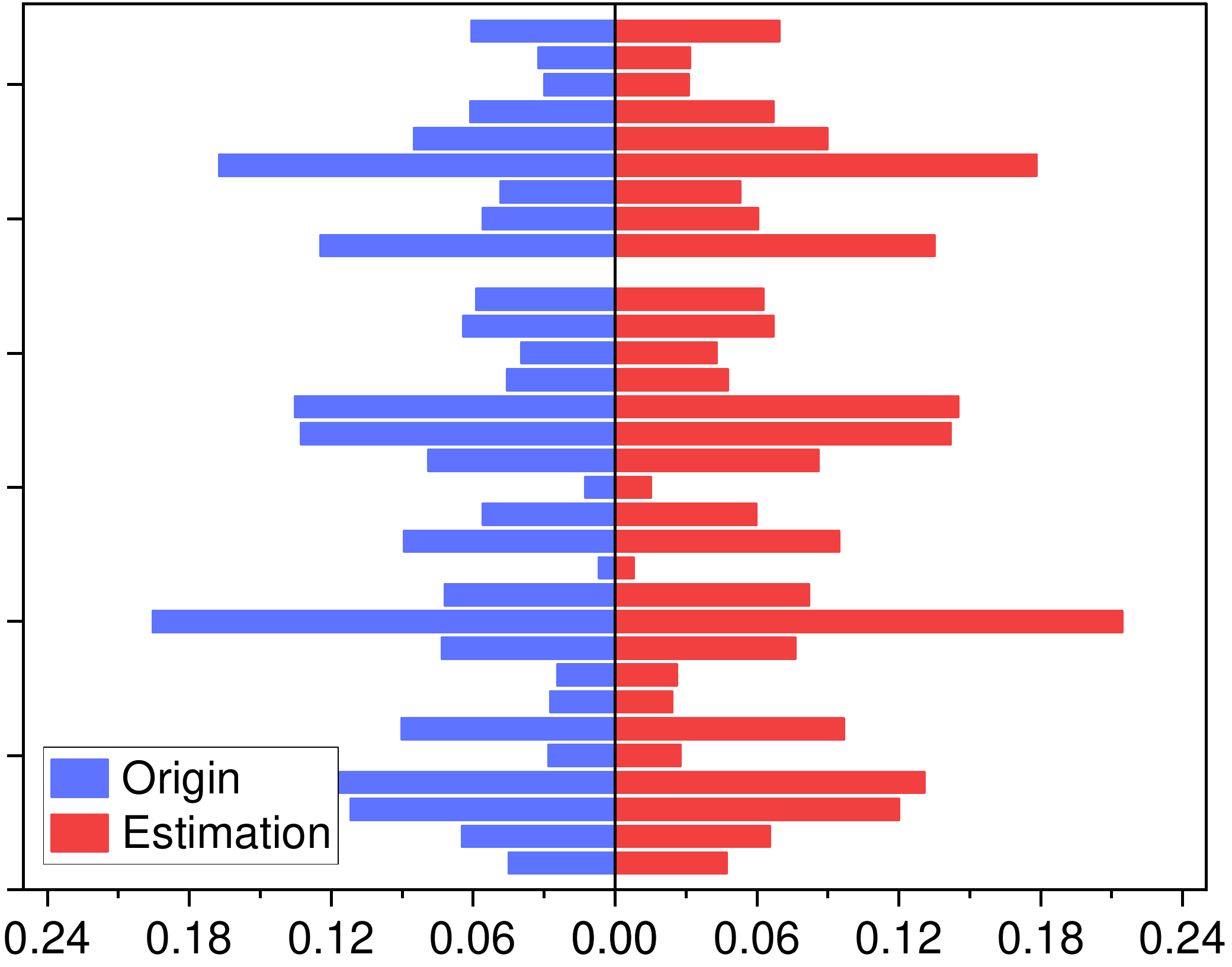}}
\end{minipage}
\caption{Comparison of the original and estimated distributions. For each subﬁgure, the horizontal axis is the expectation, and the vertical axis represents different features. The red indicates the distribution estimation, and the blue is the original distribution. More comparisons are provided in the supplementary.}
\label{fig:ablation-2}
\end{figure}
\begin{figure}[t]
\centering
\begin{minipage}[!t]{1.\linewidth}
\centering
\subfigure[layer-1]{
	\centering
	\includegraphics[scale=0.17]{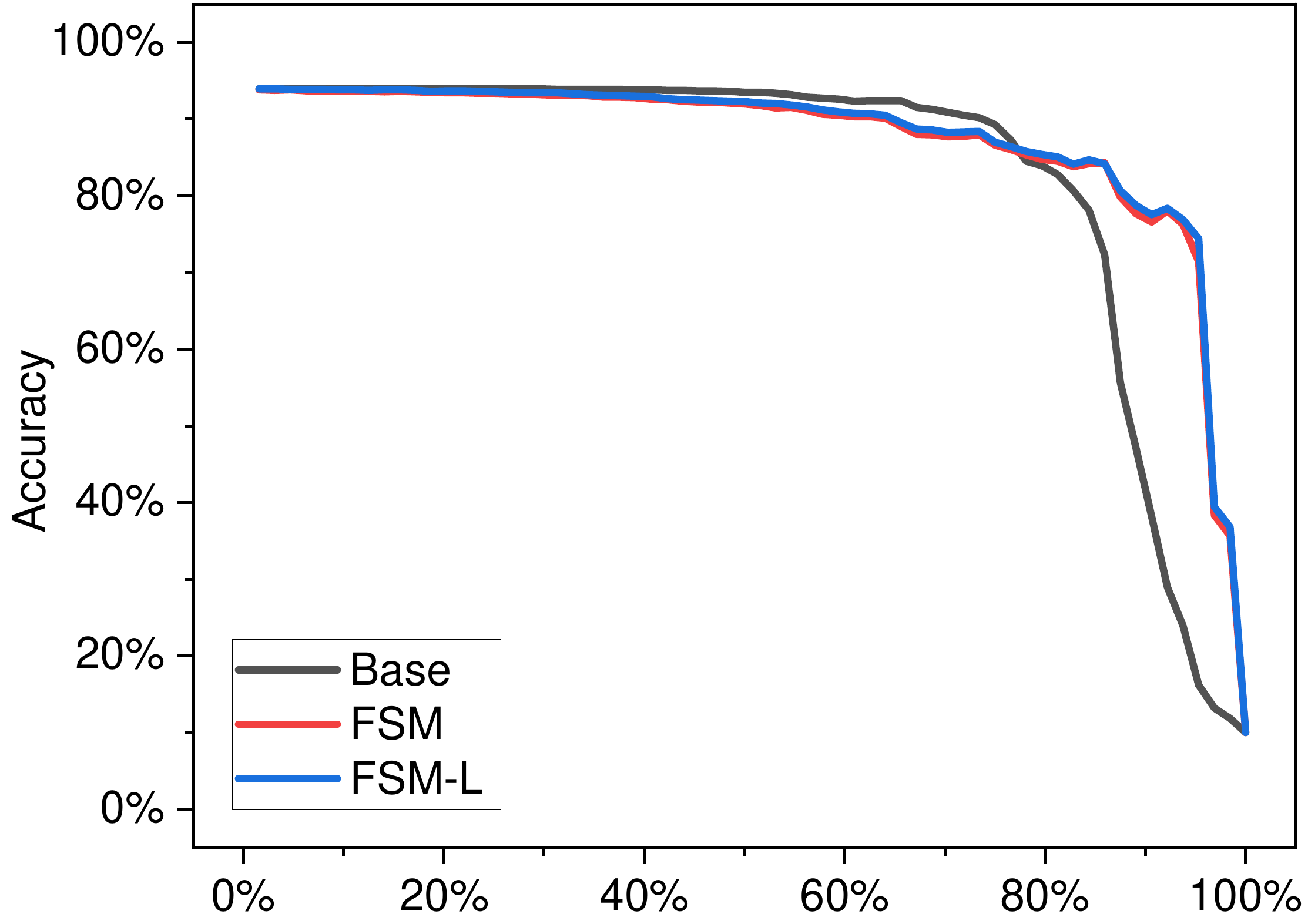}}
\subfigure[layer-6]{
	\centering
	\includegraphics[scale=0.17]{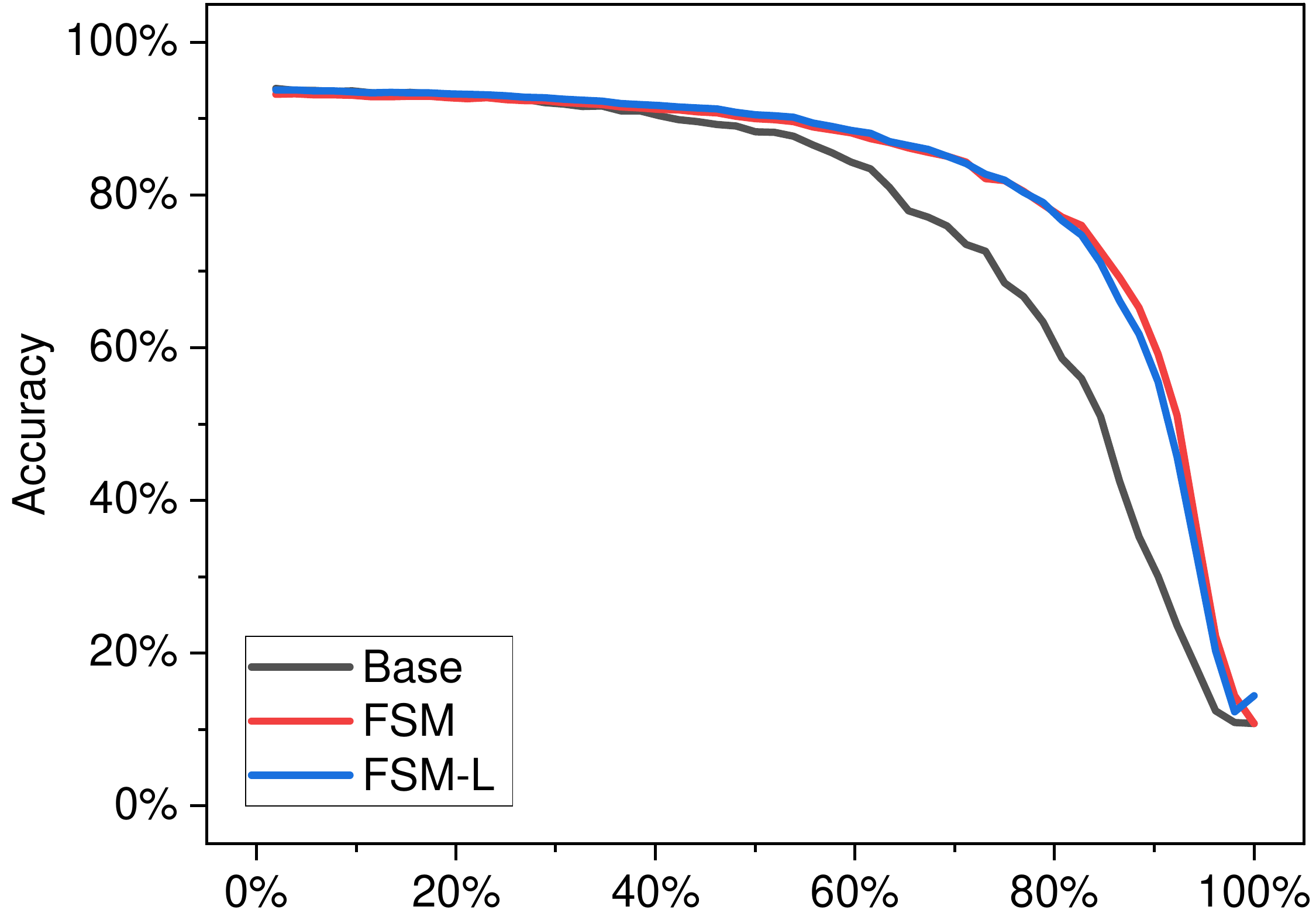}}
\subfigure[layer-11]{
	\centering
	\includegraphics[scale=0.17]{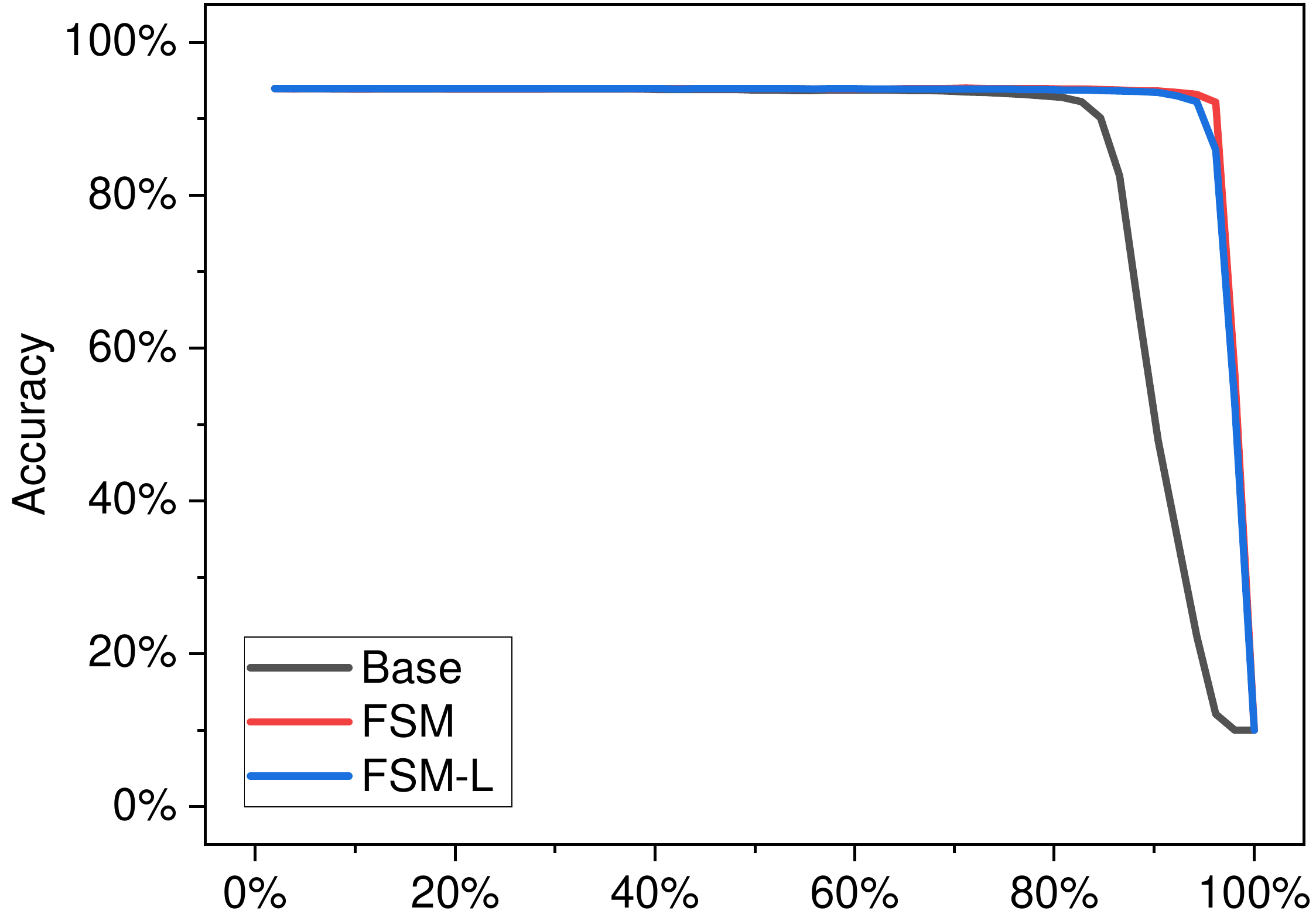}}
\end{minipage}
\caption{The effect of evaluation error $\lambda$ for channel pruning on VGGNet. For each subfigure, the FSM-L means that the expectation evaluation values are amended with $\lambda$.}
\label{fig:ablation-3}
\end{figure}
\begin{figure}[t]
\centering
\begin{minipage}[!t]{1.\linewidth}
\centering
\subfigure[layer-1]{
	\centering
	\includegraphics[scale=0.17]{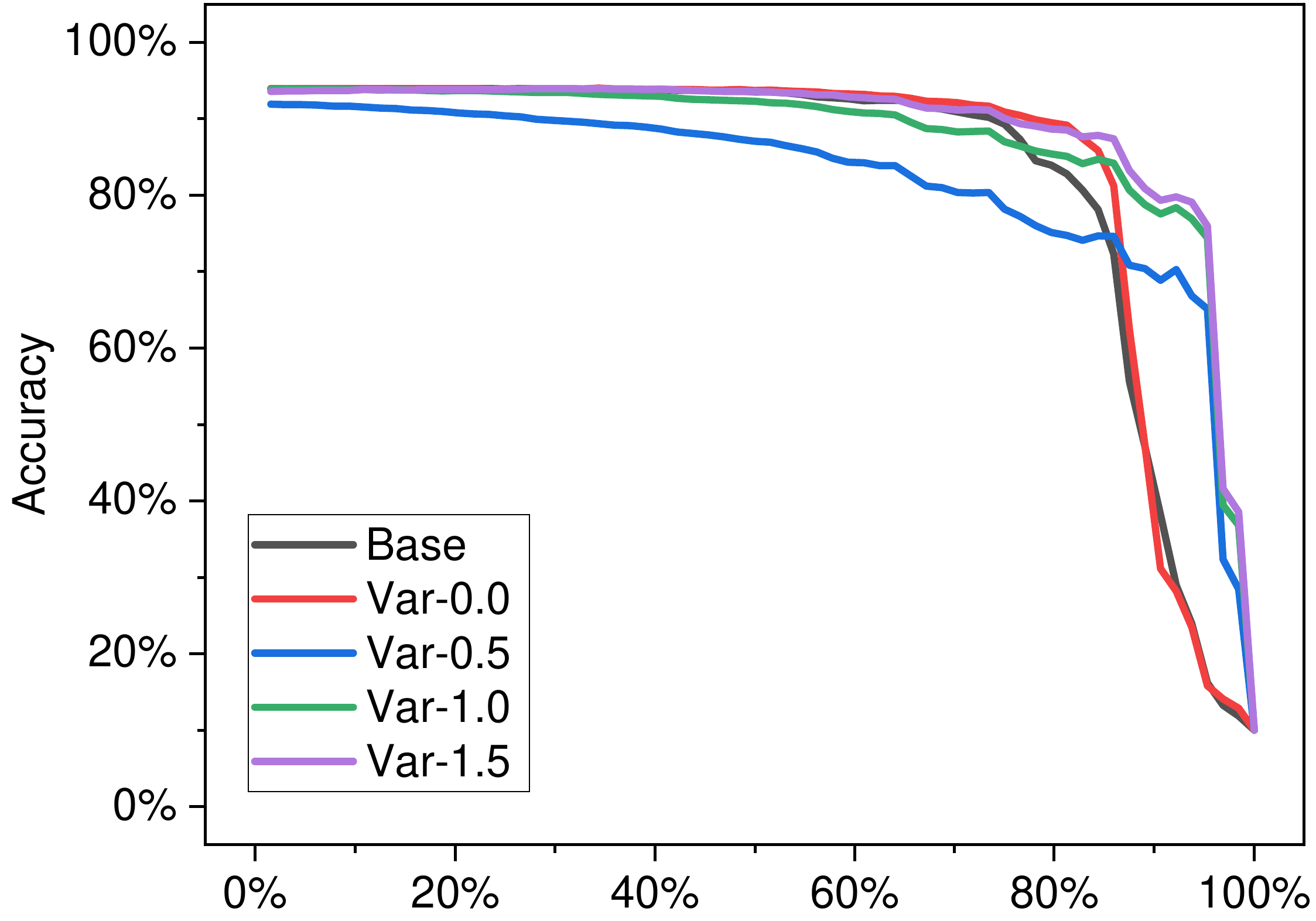}}
\subfigure[layer-6]{
	\centering
	\includegraphics[scale=0.17]{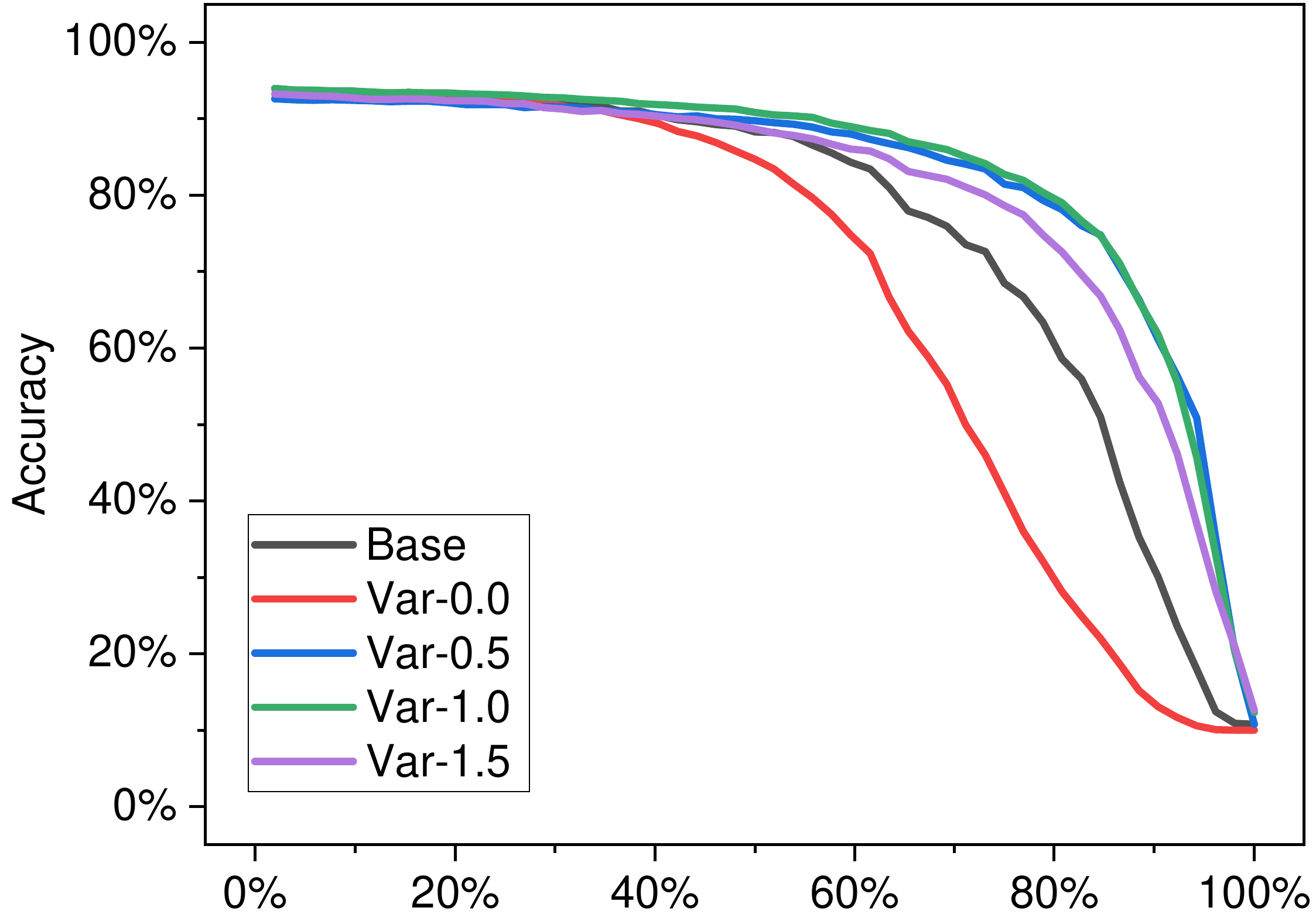}}
\subfigure[layer-11]{
	\centering
	\includegraphics[scale=0.17]{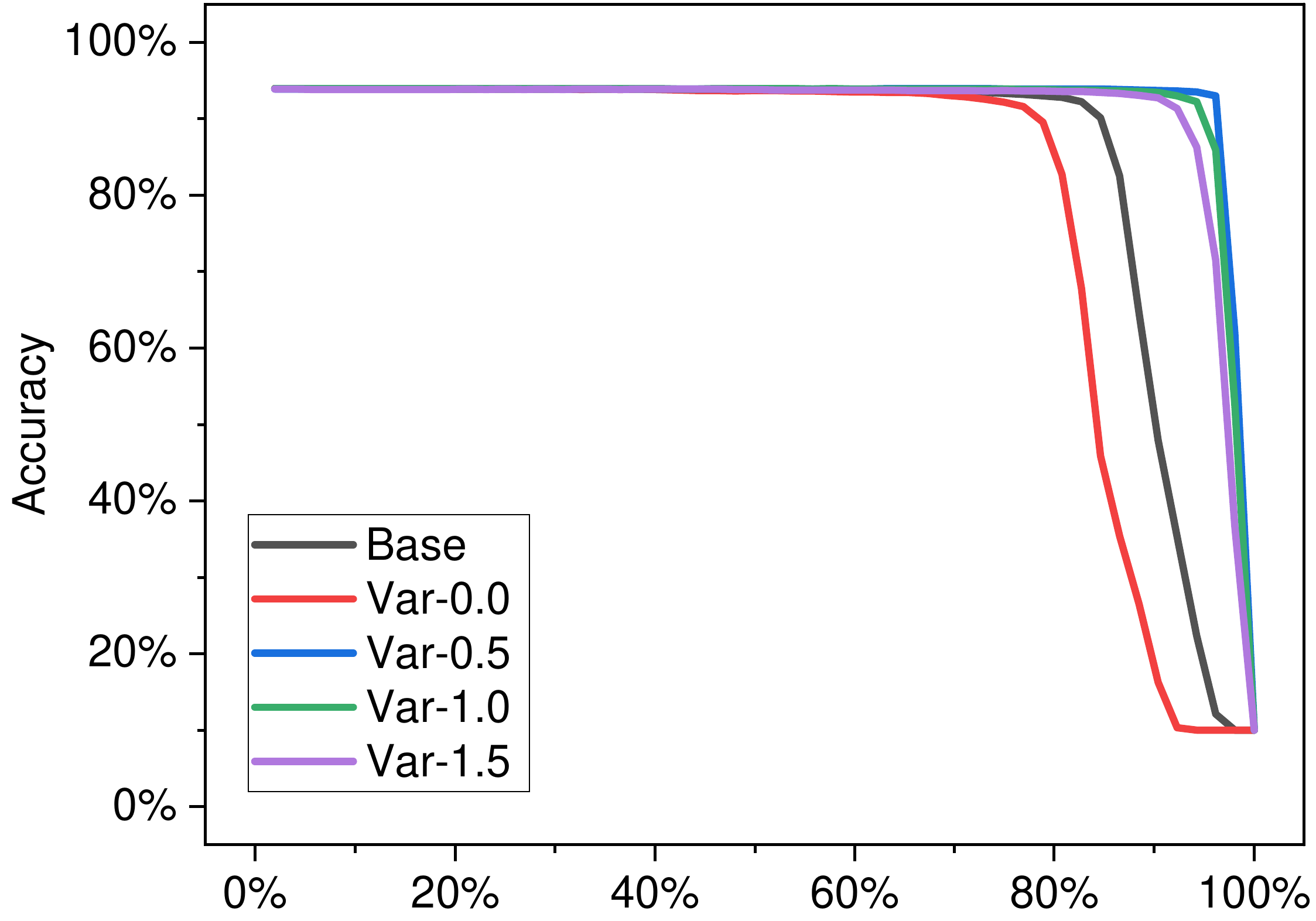}}
\end{minipage}
\caption{Comparisons of the variance estimation with different magnitude coefficients on VGGNet. Names such as "Var-1.5" stand for reduce the variance $Var[x^{(k)}_{i}]$ by $(1.5\times \frac{\widehat{d}_i}{d_i})\%$.
Similarly, the horizontal axis represents the compression rate, and the vertical axis is the accuracy.}
\label{fig:ablation-4}
\end{figure}
The error of the feature distribution evaluation is shown in Fig.\;\ref{fig:ablation-2}.
The differences between the proposed evaluation method and the true distribution are compared at different layer depths.
It can be seen that the evaluation error is in an acceptable range.
In addition, our experiments show that expectation and layer depth are negatively correlated.
For very small expectation values, our method can still estimate them accurately.
\subsection{Effect of the Evaluation Error $\lambda$}

In Section\;\ref{Distribution Optimization}, we present the estimation error $\lambda$ to rectify the evaluated distribution. 
As shown in Fig.\;\ref{fig:ablation-3}, the method using $\lambda$  generates a better performance at low compression rates.
Also, no additional accuracy loss is caused at high compression rates.
\subsection{Effect of the Variance Adjustment Coefficients}

In Section\;\ref{Distribution Optimization}, we empirically reduce the variance $Var[x^{(k)}_{i}]$ by $\frac{\widehat{d}_i}{d_i} $\% because it is difficult to calculate it directly.
As shown in Fig.\;\ref{fig:ablation-4}, we adopt different magnitude coefficients to adjust the variances of pruned models.
The results show that the VAR-1.0, adopted in our work, achieves the best performance in almost all cases.

\section{Conclusions}
In this paper, we propose a novel channel pruning method by minimizing feature shift.
We first prove the existence of feature shift mathematically, inspired by the investigation of accuracy curves in channel pruning.
The feature shift is used to explain why the accuracy plummets when the compression rate exceeds a critical value.
Based on this, an efficient channel pruning method (FSM) is proposed, which performs well, especially at high compression rates.
Then, we present a feature shift evaluation method that does not traverse the training data set.
In addition, a distribution optimization method is designed to improve the efficiency of model compression and is plug-and-play.
Extensive experiments and rigorous ablation studies demonstrate the effectiveness of the proposed FSM for channel pruning.
In the future, we will further research the impact of multi-branch architecture on the FSM.
Also, the combination with other pruning methods is worth trying.

\subsubsection{Acknowledgements}
The work was supported by National Natural Science Foundation of China (Grant Nos. 61976246 and U20A20227), Natural Science Foundation of Chongqing (Grant No. cstc2020jcyj-msxm X0385)

\bibliographystyle{splncs}
\bibliography{egbib}

\begin{thebibliography}{10}

\bibitem{He2016DeepRL}
He, K., Zhang, X., Ren, S., Sun, J.:
\newblock Deep residual learning for image recognition.
\newblock 2016 IEEE Conference on Computer Vision and Pattern Recognition
  (CVPR) (2016)  770--778

\bibitem{Huang2017DenselyCC}
Huang, G., Liu, Z., Weinberger, K.Q.:
\newblock Densely connected convolutional networks.
\newblock 2017 IEEE Conference on Computer Vision and Pattern Recognition
  (CVPR) (2017)  2261--2269

\bibitem{Krizhevsky2012ImageNetCW}
Krizhevsky, A., Sutskever, I., Hinton, G.E.:
\newblock Imagenet classification with deep convolutional neural networks.
\newblock Communications of the ACM \textbf{60} (2012)  84 -- 90

\bibitem{Girdhar2019DistInitLV}
Girdhar, R., Tran, D., Torresani, L., Ramanan, D.:
\newblock Distinit: Learning video representations without a single labeled
  video.
\newblock 2019 IEEE/CVF International Conference on Computer Vision (ICCV)
  (2019)  852--861

\bibitem{Jiang2018DeepVSAD}
Jiang, L., Xu, M., Liu, T., Qiao, M., Wang, Z.:
\newblock Deepvs: A deep learning based video saliency prediction approach.
\newblock In: ECCV. (2018)

\bibitem{Lin2019BMNBN}
Lin, T., Liu, X., Li, X., Ding, E., Wen, S.:
\newblock Bmn: Boundary-matching network for temporal action proposal
  generation.
\newblock 2019 IEEE/CVF International Conference on Computer Vision (ICCV)
  (2019)  3888--3897

\bibitem{Girshick2014RichFH}
Girshick, R.B., Donahue, J., Darrell, T., Malik, J.:
\newblock Rich feature hierarchies for accurate object detection and semantic
  segmentation.
\newblock 2014 IEEE Conference on Computer Vision and Pattern Recognition
  (2014)  580--587

\bibitem{Guo2020HitDetectorHT}
Guo, J., Han, K., Wang, Y., Zhang, C., Yang, Z., Wu, H., Chen, X., Xu, C.:
\newblock Hit-detector: Hierarchical trinity architecture search for object
  detection.
\newblock 2020 IEEE/CVF Conference on Computer Vision and Pattern Recognition
  (CVPR) (2020)  11402--11411

\bibitem{Ren2015FasterRT}
Ren, S., He, K., Girshick, R.B., Sun, J.:
\newblock Faster r-cnn: Towards real-time object detection with region proposal
  networks.
\newblock IEEE Transactions on Pattern Analysis and Machine Intelligence
  \textbf{39} (2015)  1137--1149

\bibitem{Chen2018DeepLabSI}
Chen, L.C., Papandreou, G., Kokkinos, I., Murphy, K.P., Yuille, A.L.:
\newblock Deeplab: Semantic image segmentation with deep convolutional nets,
  atrous convolution, and fully connected crfs.
\newblock IEEE Transactions on Pattern Analysis and Machine Intelligence
  \textbf{40} (2018)  834--848

\bibitem{Chen2018EncoderDecoderWA}
Chen, L.C., Zhu, Y., Papandreou, G., Schroff, F., Adam, H.:
\newblock Encoder-decoder with atrous separable convolution for semantic image
  segmentation.
\newblock ArXiv \textbf{abs/1802.02611} (2018)

\bibitem{Shelhamer2017FullyCN}
Shelhamer, E., Long, J., Darrell, T.:
\newblock Fully convolutional networks for semantic segmentation.
\newblock IEEE Transactions on Pattern Analysis and Machine Intelligence
  \textbf{39} (2017)  640--651

\bibitem{Raja2019ObtainingSI}
Raja, K.B., Raghavendra, R., Busch, C.:
\newblock Obtaining stable iris codes exploiting low-rank tensor space and
  spatial structure aware refinement for better iris recognition.
\newblock 2019 International Conference on Biometrics (ICB) (2019)  1--8

\bibitem{Liu2020ReActNetTP}
Liu, Z., Shen, Z., Savvides, M., Cheng, K.T.:
\newblock Reactnet: Towards precise binary neural network with generalized
  activation functions.
\newblock ArXiv \textbf{abs/2003.03488} (2020)

\bibitem{Han2015LearningBW}
Han, S., Pool, J., Tran, J., Dally, W.J.:
\newblock Learning both weights and connections for efficient neural network.
\newblock ArXiv \textbf{abs/1506.02626} (2015)

\bibitem{Li2017PruningFF}
Li, H., Kadav, A., Durdanovic, I., Samet, H., Graf, H.P.:
\newblock Pruning filters for efficient convnets.
\newblock ArXiv \textbf{abs/1608.08710} (2017)

\bibitem{Lin2020HRankFP}
Lin, M., Ji, R., Wang, Y., Zhang, Y., Zhang, B., Tian, Y., Shao, L.:
\newblock Hrank: Filter pruning using high-rank feature map.
\newblock 2020 IEEE/CVF Conference on Computer Vision and Pattern Recognition
  (CVPR) (2020)  1526--1535

\bibitem{Hinton2015DistillingTK}
Hinton, G.E., Vinyals, O., Dean, J.:
\newblock Distilling the knowledge in a neural network.
\newblock ArXiv \textbf{abs/1503.02531} (2015)

\bibitem{Howard2017MobileNetsEC}
Howard, A.G., Zhu, M., Chen, B., Kalenichenko, D., Wang, W., Weyand, T.,
  Andreetto, M., Adam, H.:
\newblock Mobilenets: Efficient convolutional neural networks for mobile vision
  applications.
\newblock ArXiv \textbf{abs/1704.04861} (2017)

\bibitem{He2018PruningFV}
He, Y., Liu, P., Wang, Z., Yang, Y.:
\newblock Pruning filter via geometric median for deep convolutional neural
  networks acceleration.
\newblock ArXiv \textbf{abs/1811.00250} (2018)

\bibitem{Luo2017ThiNetAF}
Luo, J.H., Wu, J., Lin, W.:
\newblock Thinet: A filter level pruning method for deep neural network
  compression.
\newblock 2017 IEEE International Conference on Computer Vision (ICCV) (2017)
  5068--5076

\bibitem{Li2021TowardsCC}
Li, Y., Lin, S., Liu, J., Ye, Q., Wang, M., Chao, F., Yang, F., Ma, J., Tian,
  Q., Ji, R.:
\newblock Towards compact cnns via collaborative compression.
\newblock 2021 IEEE/CVF Conference on Computer Vision and Pattern Recognition
  (CVPR) (2021)  6434--6443

\bibitem{Wang2021ConvolutionalNN}
Wang, Z., Li, C., Wang, X.:
\newblock Convolutional neural network pruning with structural redundancy
  reduction.
\newblock 2021 IEEE/CVF Conference on Computer Vision and Pattern Recognition
  (CVPR) (2021)  14908--14917

\bibitem{Chin2020TowardsEM}
Chin, T.W., Ding, R., Zhang, C., Marculescu, D.:
\newblock Towards efficient model compression via learned global ranking.
\newblock 2020 IEEE/CVF Conference on Computer Vision and Pattern Recognition
  (CVPR) (2020)  1515--1525

\bibitem{Liu2017LearningEC}
Liu, Z., Li, J., Shen, Z., Huang, G., Yan, S., Zhang, C.:
\newblock Learning efficient convolutional networks through network slimming.
\newblock 2017 IEEE International Conference on Computer Vision (ICCV) (2017)
  2755--2763

\bibitem{He2017ChannelPF}
He, Y., Zhang, X., Sun, J.:
\newblock Channel pruning for accelerating very deep neural networks.
\newblock 2017 IEEE International Conference on Computer Vision (ICCV) (2017)
  1398--1406

\bibitem{Singh2020LeveragingFC}
Singh, P., Verma, V.K., Rai, P., Namboodiri, V.P.:
\newblock Leveraging filter correlations for deep model compression.
\newblock 2020 IEEE Winter Conference on Applications of Computer Vision (WACV)
  (2020)  824--833

\bibitem{Wang2021ModelPB}
Wang, Z., jun Liu, X., Huang, L., Chen, Y., Zhang, Y., Lin, Z., Wang, R.:
\newblock Model pruning based on quantified similarity of feature maps.
\newblock ArXiv \textbf{abs/2105.06052} (2021)

\bibitem{He2020LearningFP}
He, Y., Ding, Y., Liu, P., Zhu, L., Zhang, H., Yang, Y.:
\newblock Learning filter pruning criteria for deep convolutional neural
  networks acceleration.
\newblock 2020 IEEE/CVF Conference on Computer Vision and Pattern Recognition
  (CVPR) (2020)  2006--2015

\bibitem{He2018SoftFP}
He, Y., Kang, G., Dong, X., Fu, Y., Yang, Y.:
\newblock Soft filter pruning for accelerating deep convolutional neural
  networks.
\newblock ArXiv \textbf{abs/1808.06866} (2018)

\bibitem{Huang2018DataDrivenSS}
Huang, Z., Wang, N.:
\newblock Data-driven sparse structure selection for deep neural networks.
\newblock ArXiv \textbf{abs/1707.01213} (2018)

\bibitem{Zhuang2018DiscriminationawareCP}
Zhuang, Z., Tan, M., Zhuang, B., Liu, J., Guo, Y., Wu, Q., Huang, J., Zhu,
  J.H.:
\newblock Discrimination-aware channel pruning for deep neural networks.
\newblock In: NeurIPS. (2018)

\bibitem{Ioffe2015BatchNA}
Ioffe, S., Szegedy, C.:
\newblock Batch normalization: Accelerating deep network training by reducing
  internal covariate shift.
\newblock ArXiv \textbf{abs/1502.03167} (2015)

\bibitem{Glorot2011DeepSR}
Glorot, X., Bordes, A., Bengio, Y.:
\newblock Deep sparse rectifier neural networks.
\newblock In: AISTATS. (2011)

\bibitem{Krizhevsky2009LearningML}
Krizhevsky, A.:
\newblock Learning multiple layers of features from tiny images.
\newblock (2009)

\bibitem{Szegedy2015GoingDW}
Szegedy, C., Liu, W., Jia, Y., Sermanet, P., Reed, S.E., Anguelov, D., Erhan,
  D., Vanhoucke, V., Rabinovich, A.:
\newblock Going deeper with convolutions.
\newblock 2015 IEEE Conference on Computer Vision and Pattern Recognition
  (CVPR) (2015)  1--9

\bibitem{Sandler2018MobileNetV2IR}
Sandler, M., Howard, A.G., Zhu, M., Zhmoginov, A., Chen, L.C.:
\newblock Mobilenetv2: Inverted residuals and linear bottlenecks.
\newblock 2018 IEEE/CVF Conference on Computer Vision and Pattern Recognition
  (2018)  4510--4520

\bibitem{Simonyan2015VeryDC}
Simonyan, K., Zisserman, A.:
\newblock Very deep convolutional networks for large-scale image recognition.
\newblock CoRR \textbf{abs/1409.1556} (2015)

\bibitem{Paszke2017AutomaticDI}
Paszke, A., Gross, S., Chintala, S., Chanan, G., Yang, E., DeVito, Z., Lin, Z.,
  Desmaison, A., Antiga, L., Lerer, A.:
\newblock Automatic differentiation in pytorch.
\newblock (2017)

\bibitem{Peng2019CollaborativeCP}
Peng, H., Wu, J., Chen, S., Huang, J.:
\newblock Collaborative channel pruning for deep networks.
\newblock In: ICML. (2019)

\bibitem{Gao2021NetworkPV}
Gao, S., Huang, F., Cai, W.T., Huang, H.:
\newblock Network pruning via performance maximization.
\newblock 2021 IEEE/CVF Conference on Computer Vision and Pattern Recognition
  (CVPR) (2021)  9266--9276

\bibitem{Li2020DHPDM}
Li, Y., Gu, S., Zhang, K., Gool, L.V., Timofte, R.:
\newblock Dhp: Differentiable meta pruning via hypernetworks.
\newblock ArXiv \textbf{abs/2003.13683} (2020)

\bibitem{Kang2020OperationAwareSC}
Kang, M., Han, B.:
\newblock Operation-aware soft channel pruning using differentiable masks.
\newblock In: ICML. (2020)

\bibitem{Lin2019TowardsOS}
Lin, S., Ji, R., Yan, C., Zhang, B., Cao, L., Ye, Q., Huang, F., Doermann,
  D.S.:
\newblock Towards optimal structured cnn pruning via generative adversarial
  learning.
\newblock 2019 IEEE/CVF Conference on Computer Vision and Pattern Recognition
  (CVPR) (2019)  2785--2794

\bibitem{Zhang2021ExplorationAE}
Zhang, Y., Gao, S., Huang, H.:
\newblock Exploration and estimation for model compression.
\newblock 2021 IEEE/CVF International Conference on Computer Vision (ICCV)
  (2021)  477--486

\bibitem{Liu2019MetaPruningML}
Liu, Z., Mu, H., Zhang, X., Guo, Z., Yang, X., Cheng, K., Sun, J.:
\newblock Metapruning: Meta learning for automatic neural network channel
  pruning.
\newblock 2019 IEEE/CVF International Conference on Computer Vision (ICCV)
  (2019)  3295--3304

\bibitem{You2019GateDG}
You, Z., Yan, K., Ye, J., Ma, M., Wang, P.:
\newblock Gate decorator: Global filter pruning method for accelerating deep
  convolutional neural networks.
\newblock ArXiv \textbf{abs/1909.08174} (2019)

\bibitem{Oh2022BatchNT}
Oh, J., Kim, H., Baik, S., Hong, C., Lee, K.M.:
\newblock Batch normalization tells you which filter is important.
\newblock 2022 IEEE/CVF Winter Conference on Applications of Computer Vision
  (WACV) (2022)  3351--3360

\bibitem{wang2020neural}
Wang, H., Qin, C., Zhang, Y., Fu, Y.:
\newblock Neural pruning via growing regularization.
\newblock In: International Conference on Learning Representations. (2020)

\bibitem{wang2021convolutional}
Wang, Z., Li, C., Wang, X.:
\newblock Convolutional neural network pruning with structural redundancy
  reduction.
\newblock In: Proceedings of the IEEE/CVF Conference on Computer Vision and
  Pattern Recognition. (2021)  14913--14922

\bibitem{He2015DelvingDI}
He, K., Zhang, X., Ren, S., Sun, J.:
\newblock Delving deep into rectifiers: Surpassing human-level performance on
  imagenet classification.
\newblock 2015 IEEE International Conference on Computer Vision (ICCV) (2015)
  1026--1034

\end{thebibliography}

\end{document}